\documentclass{ifm-base}

\usepackage{amsmath,amsfonts,amssymb,mathtools,bm}  
\usepackage{thmtools}          
\usepackage{algorithm}         
\usepackage{algpseudocode}     
\usepackage[shortlabels]{enumitem}          
\usepackage{siunitx}           
\usepackage{tikz}              
\usepackage{booktabs}          
\usepackage{subcaption}        

\usepackage{hyperref}       
\usepackage{url}            

\usepackage{xurl}          
\usepackage{booktabs}       
\usepackage{amsfonts}       
\usepackage{nicefrac}       
\usepackage{microtype}      
\usepackage{graphicx}
\usepackage{subcaption} 
\usepackage{xspace}
\usepackage[frozencache,cachedir=.]{minted}
\usepackage{longtable}

\usepackage{nicematrix}
\usepackage{multirow}
\usepackage{makecell}
\usepackage{soul}  
\usepackage{wrapfig}
\usepackage{enumitem}

\usepackage{colortbl}
\usepackage{ulem}

\usepackage{xspace}
\newcommand{\huggingface}{\raisebox{-1.5pt}{\includegraphics[height=1.05em]{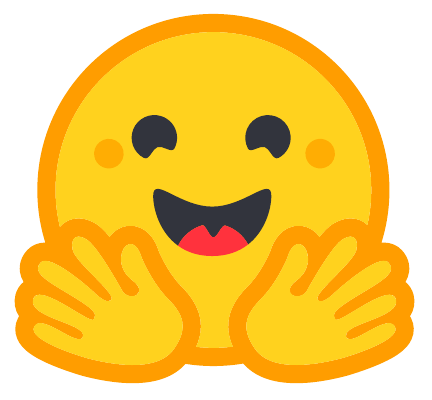}}\xspace}
\newcommand{\github}{\raisebox{-1.5pt}{\includegraphics[height=1.05em]{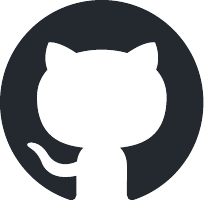}}\xspace}
\newcommand{\logo}{\raisebox{-1.5pt}{\includegraphics[height=1.05em]{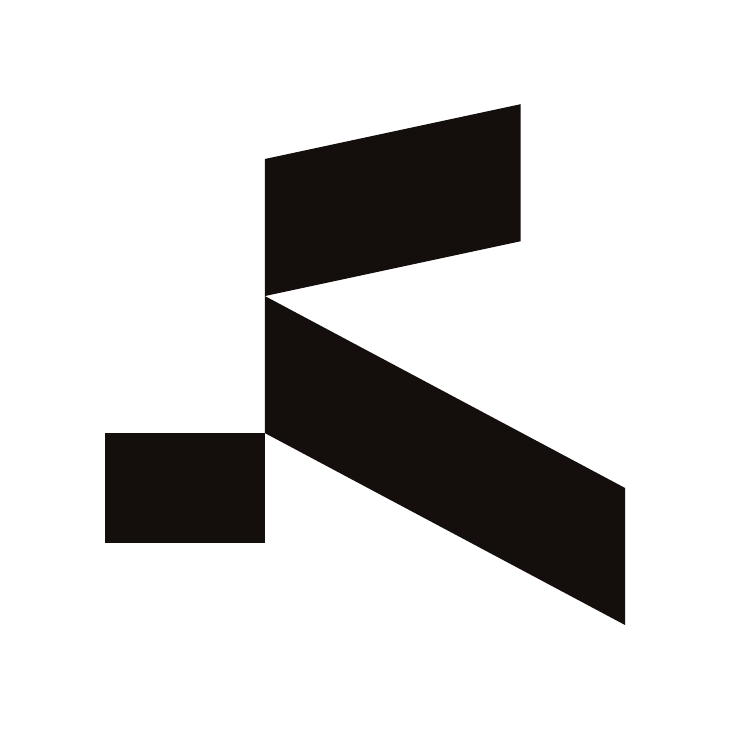}}\xspace}

\newcommand\ours{\textsc{K2-Think}\xspace}

\definecolor{cellHighlight}{HTML}{dbefff}

\newcommand\rurl[1]{%
    \href{https://#1}{\nolinkurl{#1}}%
}

\title{{\fontsize{18}{16}\selectfont \bfseries K2-Think: A Parameter-Efficient Reasoning System}}

\author[*]{Zhoujun Cheng}\authorsep{,}
\author[*]{Richard Fan}\authorsep{,}
\author[*]{Shibo Hao}\authorsep{,}
\author[*,O]{Taylor W. Killian}\authorsep{,}
\author[*]{Haonan Li}\authorsep{,}
\author[*]{Suqi Sun}\authorsep{\\} 
\author[]{Hector Ren}\authorsep{,}
\author[]{Alexander Moreno}\authorsep{,}
\author[]{Daqian Zhang}\authorsep{,}
\author[]{Tianjun Zhong}\authorsep{,} 
\author[]{Yuxin Xiong}\authorsep{,}
\author[]{Yuanzhe Hu}\authorsep{,}
\author[]{Yutao Xie}\authorsep{\\}
\author[]{Xudong Han}\authorsep{,}
\author[]{Yuqi Wang}\authorsep{,}
\author[]{Varad Pimpalkhute}\authorsep{,}
\author[]{Yonghao Zhuang}\authorsep{,}
\author[]{Aaryamonvikram Singh}\authorsep{,}
\author[]{Xuezhi Liang}\authorsep{\\}
\author[]{Anze Xie}\authorsep{,}
\author[]{Jianshu She}\authorsep{,}
\author[]{Desai Fan}\authorsep{,}
\author[]{Chengqian Gao}\authorsep{,}
\author[]{Liqun Ma}\authorsep{,}
\author[]{Mikhail Yurochkin}\authorsep{,}
\author[]{John Maggs}\authorsep{\\}
\author[]{Xuezhe Ma}\authorsep{,}
\author[]{Guowei He}\authorsep{,}
\author[]{Zhiting Hu}\authorsep{,}
\author[*,O]{Zhengzhong Liu}\authorsep{,}
\author[O]{Eric P. Xing}\authorsep{}

\affiliation[]{Institute of Foundation Models, Mohamed bin Zayed University of Artificial Intelligence}

\contribution[*]{Core Contributors (listed alphabetically)}
\contribution[O]{Corresponding Authors}

\abstract{
We introduce \ours{}, a reasoning system that achieves frontier performance with just a 32B parameter model — surpassing or matching much larger models such as GPT-OSS 120B and DeepSeek v3.1. Built on the Qwen2.5 base model, our system demonstrates that smaller models can compete at the highest levels through synergistic combination of advanced post-training and test-time computation techniques.
Our approach is built on top of six key technical pillars: Long Chain-of-thought Supervised Finetuning, Reinforcement Learning with Verifiable Rewards (RLVR), Agentic planning prior to reasoning, Test-time Scaling, Speculative Decoding, and Inference-optimized Hardware, using only publicly available open-source datasets. \ours{} prioritizes mathematical reasoning, achieving state-of-the-art scores on public benchmarks for open source models, while also maintaining strong performance on other domains such as Code and Science. 
Our results validate that a more parameter-efficient model like K2-Think 32B can rival state-of-the-art systems through an integrative post-train recipe including long chain-of-thought training and strategic inference-time enhancements, paving the way for more accessible and affordable open-source reasoning systems. We have made \ours{} freely available at \rurl{k2think.ai} demonstrating best-in-class inference speeds, through the Cerebras Wafer-Scale Engine, delivering upwards of 2,000 tokens per second per request.
}

\begin{document}
\maketitle

\footnotetext[\value{footnote}]{Correspondence to: \texttt{\{Eric.Xing,Hector.Liu,Taylor.Killian\}@mbzuai.ac.ae}}


\begin{minipage}{\textwidth}
    \centering
    \includegraphics[width=5cm]{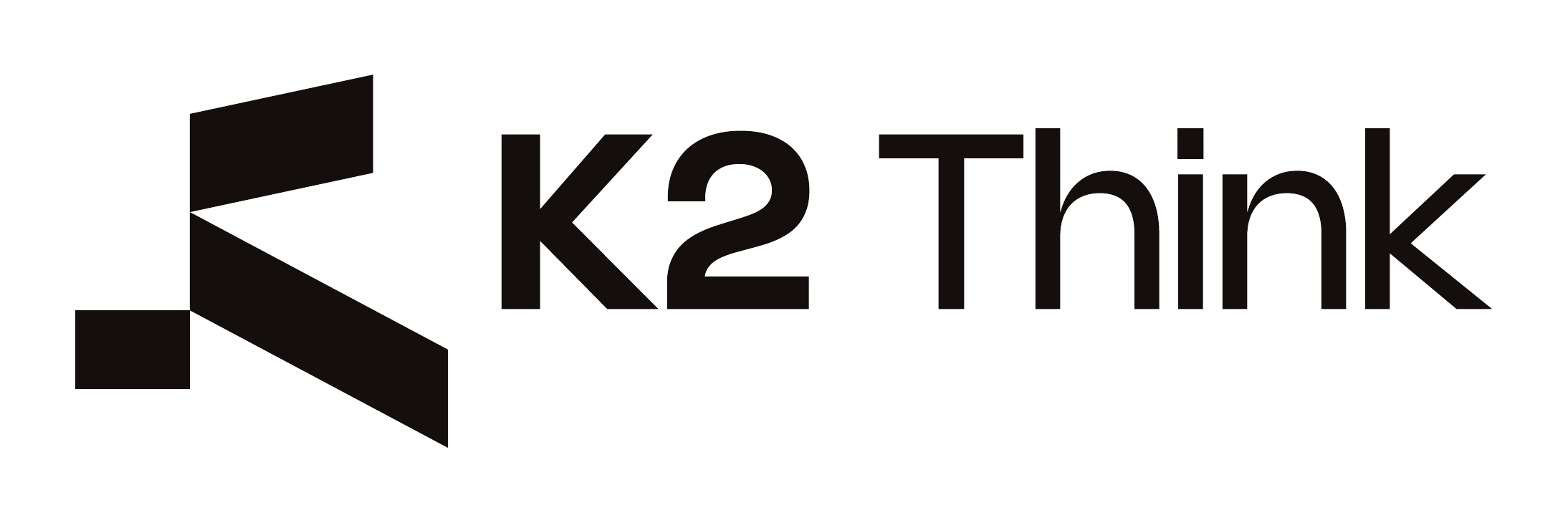}
\end{minipage}

\vspace{0.75cm}

\hfill
\begin{minipage}{0.85\textwidth}
\centering
\begin{tabular}{ l@{}l l }
\huggingface & \ours{} (Model) & \rurl{huggingface.co/LLM360/K2-Think}  \\[6pt]
\multirow{2}{*}{\github} & \multirow{2}{*}{\ours{} (Code)} & \rurl{github.com/MBZUAI-IFM/K2-Think-SFT} \\
& & \rurl{github.com/MBZUAI-IFM/K2-Think-Inference}  \\[6pt]
\logo & \ours{} (Web) & \rurl{k2think.ai} \\
\end{tabular}
\end{minipage}

\newpage

\begin{figure}
    \centering
    \includegraphics[width=0.9\linewidth]{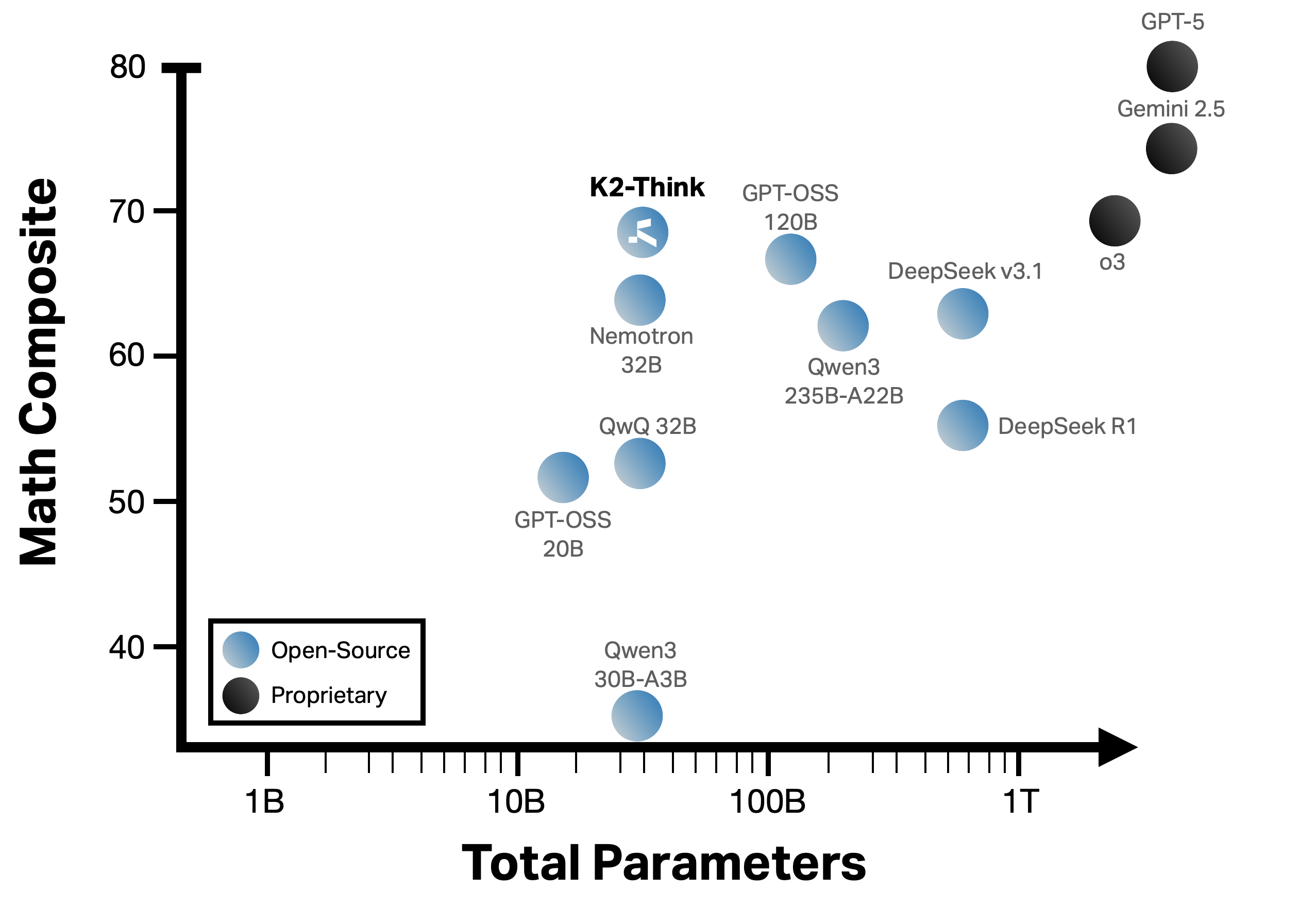}
    \caption{\ours{} exhibits remarkable parameter efficiency, providing comparable or superior performance to frontier reasoning models in complex math domains with an order of magnitude smaller model. \emph{The composite score here is the micro-average for each model over four complex math benchmarks, weighted by the number of questions in each benchmark} (AIME 2024, AIME 2025, HMMT 2025, and Omni-MATH-HARD; see Section~\ref{sec:evaluation} for the benchmark details). Note: parameter counts for proprietary models are speculative.}
    \label{fig:math_v_params}
\end{figure}

\section{Introduction}\label{sec:intro}

Recent advances in frontier reasoning models have highlighted the effectiveness of long chain-of-thought reasoning, enabled by large-scale supervised fine-tuning and reinforcement learning.
Systems like OpenAI-O3~\citep{openaio3} and Gemini 2.5~\citep{GoogleDeepMind2025Gemini} have achieved strong results on competition-level math benchmarks, complex coding tasks, and advanced scientific reasoning datasets, setting new milestones for reasoning-centered language models. These developments have also stimulated further exploration in the open-source community, where researchers have trained competitive reasoning systems with reinforcement learning~\citep{yu2025dapo,hu2025open,wang2025octothinker} and investigated mechanisms by which RL improves reasoning~\citep{zeng2025simplerl,yue2025does,shao2025spurious,min-entropy,one-shot-rl}.

\emph{In this report we introduce \ours{}}: a competitive reasoning system built from the open-weight Qwen2.5-32B base model~\citep{yang2024qwen2}. We break down our system into stages, including post-training and test-time, where an integrative recipe involving 6 major technical innovations was introduced over all stages spanning finetuning, reinforcement learning, planning, and hardware optimization to boost the base model reasoning capability, and we evaluated how each stage affects  performance. 
\textbf{These components combine to enable a model of merely 32 billion parameters, with modest test-time compute, to match the mathematical reasoning performance of proprietary frontier models}. In fact, \ours{} emerges as the top open-source model for complex math benchmarks matching or exceeding previously leading models that are orders of magnitude greater in size. Figure~\ref{fig:math_v_params} presents a plot of the global micro-average of performance (essentially dividing the total number of correct answers by the total number of questions across all test sets) of each model over four challenging math competition tasks with respect to the total number of parameters for each model. The prominent positioning of \ours{} in the top-left visually depicts its superior parameter efficiency, demonstrating that it achieves State-of-the-Art performance among open-source models with a significantly smaller total parameter count. Detailed results and discussion for the benchmarks are presented in Section~\ref{sec:evaluation}.

More specifically, \ours{} incorporates six key innovations to deliver a strong reasoning system. We first extend the base model with \textit{chain-of-thought capabilities through Supervised Fine-tuning (SFT)}, followed by \textit{Reinforcement Learning with Verifiable Rewards (RLVR)} to strengthen reasoning performance. We then enhance the model with inference-time techniques: \textit{agentic planning} and \textit{test-time scaling} using \textit{Best-of-N sampling}. Finally, we deploy \ours{} with two speed optimizations: \textit{speculative decoding} and Cerebras' Wafer-Scale Engine, an \textit{inference-optimized hardware} system. This final stage enables the model to deliver its powerful chain-of-thought reasoning capabilities with near-instantaneous response times, deployed at speeds upwards of 2000 tokens per second per user request.

With the release of \ours{}, we share our experience and make available an important advancement in open-source language modeling, that aggressive post-training engineering and test-time computation, even with a modest commodity pretrained base model, can significantly boost  reasoning capabilities in a cost-effective manner. Prior studies have reported that, in certain regimes, allocating more computation during inference can be more cost-effective than scaling model size~\citep{snell2025scaling}; for recent frontier systems—including OpenAI’s o1/o3~\citep{OpenAIo1,openaio3}, DeepSeek-R1~\citep{guo2025deepseekr1}, Google’s Gemini 2.5~\citep{GoogleDeepMind2025Gemini}, and xAI’s Grok4~\citep{xAI2025Grok4} — model capabilities have been claimed to improve with increased test-time budgets~\citep{ji2025test,yang2025towards}. 

In addition to releasing code and model weights, \textbf{we offer \ours{} through a public website and as a production-ready API endpoint.\footnote{available upon request}} This allows the community to engage directly with a living system, shifting the emphasis from static artifacts to a deployable, studyable service that can be stress-tested and iterated on in the open. As dynamic inference-time reasoning becomes more complex, our API demonstrates the requirements of sophisticated systems for top performance, and provides an operational deployment delivering robustness, safety, and efficiency under real-world constraints.

In Section~\ref{sec:k2-think} we describe the development process and deployment of \ours{}, using the Cerebras Wafer-Scale Engine. Section~\ref{sec:evaluation} presents a thorough set of evaluations and ablations that attribute gains across post-training and test-time computation. Section~\ref{sec:related_work} situates our contributions within the literature. We conclude in Section~\ref{sec:discussion} with a summary overview, discuss our motivations for deploying this model, and chart future directions for extending reasoning performance with openly released models and deployment-ready systems.

\section{\ours{} Development}\label{sec:k2-think}

We initiate \ours{}’s development to study a complete post-training recipe for enhanced reasoning and establish best practices for extending our in-house foundation models. Throughout this study, we seek to validate published best practices as well as test original test-time computation ideas.

We choose to fine-tune a 32B-scale base model for \ours{}, as:
\begin{enumerate}[(1)]
    \item it allows for fast iteration while providing strong base capabilities and
    \item its size suits both research and consumer computation frameworks.
\end{enumerate} Specifically, we selected Qwen2.5-32B as it is not tuned for reasoning, allowing us to fully validate our recipe's effectiveness.

\subsection{Phase 1: Supervised Fine Tuning}\label{sec:sft}

The initial stage of \ours{} development constitutes supervised fine-tuning (SFT) of the base model using curated long chain-of-thoughts (CoT), establishing the first pillar of our complete reasoning system. This follows the paradigm introduced by DeepSeek in the development of their R1 model~\citep{guo2025deepseekr1}. This phase of training serves to provide guidance to the pre-trained base language model for generating structured responses to complex queries. Additionally, the model is trained to adopt an expected output format in which the model's reasoning process is made clear prior to producing an answer. By providing a token-by-token supervisory signal through extended CoT, the base model's intrinsic computation capabilities are expanded substantially~\citep{wei2022chain,schuurmans2024autoregressive}.

Our SFT phase\footnote{Code, based on LLaMA-Factory, for SFT can be found at \rurl{https://github.com/MBZUAI-IFM/K2-Think-SFT}} uses the existing \texttt{AM-Thinking-v1-Distilled} dataset,\footnote{\url{https://huggingface.co/datasets/a-m-team/AM-Thinking-v1-Distilled}} composed of CoT reasoning traces and instruction/response pairs, with prompts drawn from tasks spanning mathematical reasoning, code generation, scientific reasoning, instruction following, and general chat~\citep{ji2025thinking,tian2025not}. In what follows, we will refer to this supervised fine-tuned model as \textbf{\ours{}-SFT}.

\subsubsection{Observations}\label{sec:sft_training}

\begin{figure}[t]
    \centering
    \begin{minipage}{0.49\textwidth}
        \centering
        \includegraphics[width=\linewidth]{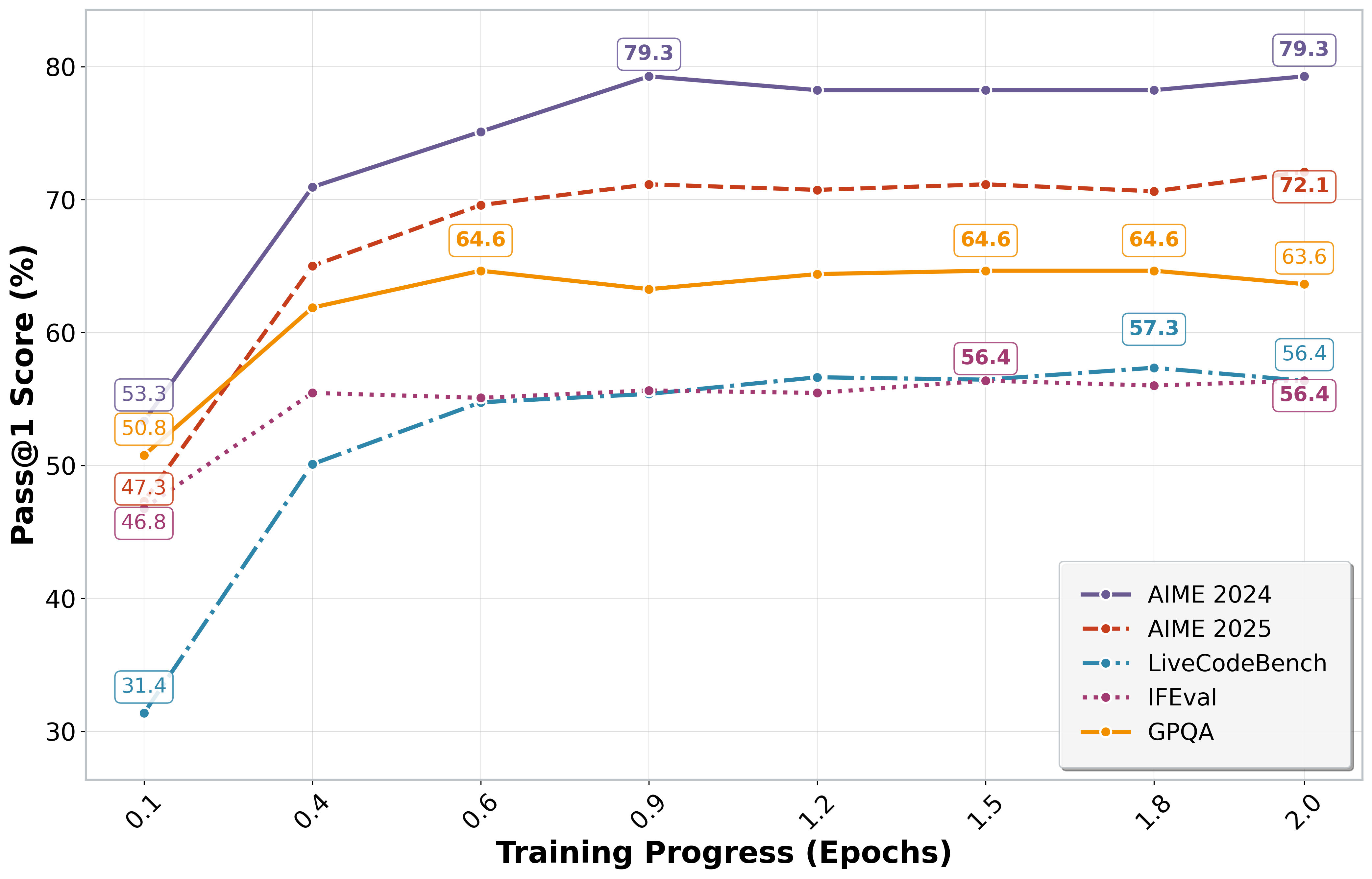}
        \caption{\textbf{Pass@1 performance over training.} Pass@1 of \ours{}-SFT across five benchmarks; the x-axis is training progress (epochs), the y-axis is pass@1 score.}
        \label{fig:amthink_step_curve}
    \end{minipage}
    \hfill
    \begin{minipage}{0.49\textwidth}
        \centering
        \includegraphics[width=0.85\linewidth]{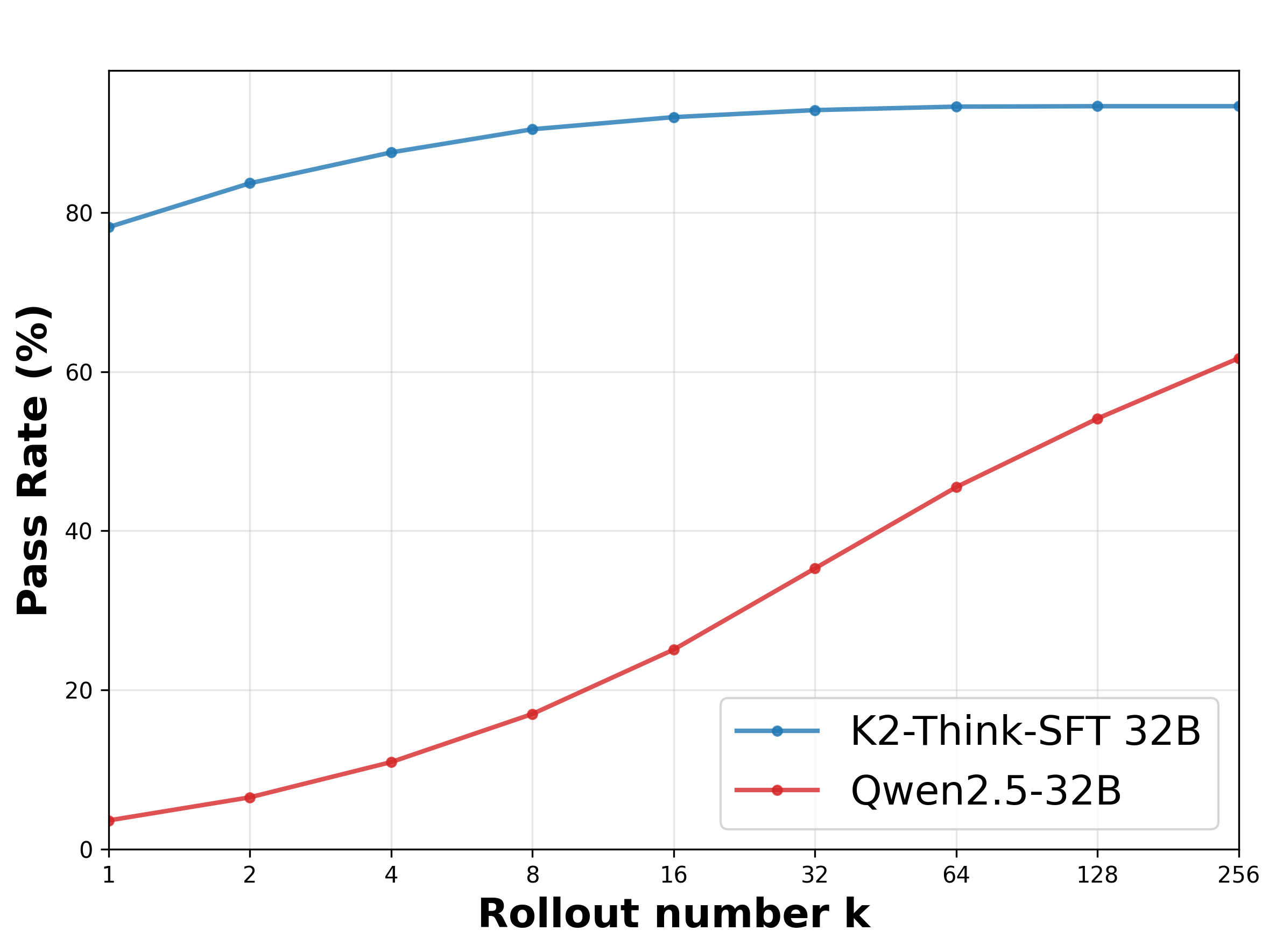}
        \caption{\textbf{Pass@k on AIME2024.} Pass@k of \ours{}-SFT and the Qwen-2.5 32B base model; the x-axis is the number of rollouts per question, the y-axis is pass rate.}
        \label{fig:sft_dataset_passk}
    \end{minipage}
\end{figure}

Our SFT experiments on Qwen2.5-32B yield several practical insights. Of particular note, we conduct a step-wise evaluation of \ours{}-SFT across five representative benchmarks. 
As shown in Figure~\ref{fig:amthink_step_curve}, performance improves rapidly within the first third of training (roughly 0.5 epoch), particularly on mathematics benchmarks (AIME 2024 and AIME 2025). After this sharp initial gain, most benchmarks plateau, with AIME 2024 stabilizing around 79.3\% pass@1 and AIME 2025 around 72.1\%. GPQA and IFEval continue to exhibit modest upward trends, while LiveCodeBench shows a slower but steady improvement up to 56.4\%. We observe that our SFT phase has reached convergence, with the model exhibiting diminishing returns to continued training on the dataset.

Apart from pass@1 scores, we also use $\text{pass@}k$ to quantify reasoning performance under a fixed sampling budget $k$. Interpreting the $\text{pass@}k$ curve as a capability boundary, we evaluate \ours{}-SFT. In Figure~\ref{fig:sft_dataset_passk}, assessing performance on AIME2024, our SFT model dominates the base model across sampling budgets. The SFT curves saturate near 93.3\% by $k\!\approx\!128$, whereas the base model continues to improve but remains well below that plateau. The growth in \ours{}-SFT performance as the sampling budget grows suggests there remains an opportunity for improvement during the following RL stage.

\subsection{Phase 2: Reinforcement Learning with Verifiable Rewards}\label{sec:rlvr}

Following the SFT stage, we perform Reinforcement Learning with Verifiable Rewards (RLVR) to train \ours{} to excel in domains with verifiable outcomes, which constitutes the second pillar of our full reasoning system. RLVR reduces the complexity and cost of preference-based alignment via RLHF~\citep{casper2023open} by directly optimizing for correctness of model generations.

\begin{figure*}[t]
    \vspace{-0.05cm}
    \centering
    \begin{subfigure}[b]{0.47\textwidth} 
        \centering
        \includegraphics[width=0.95\textwidth]{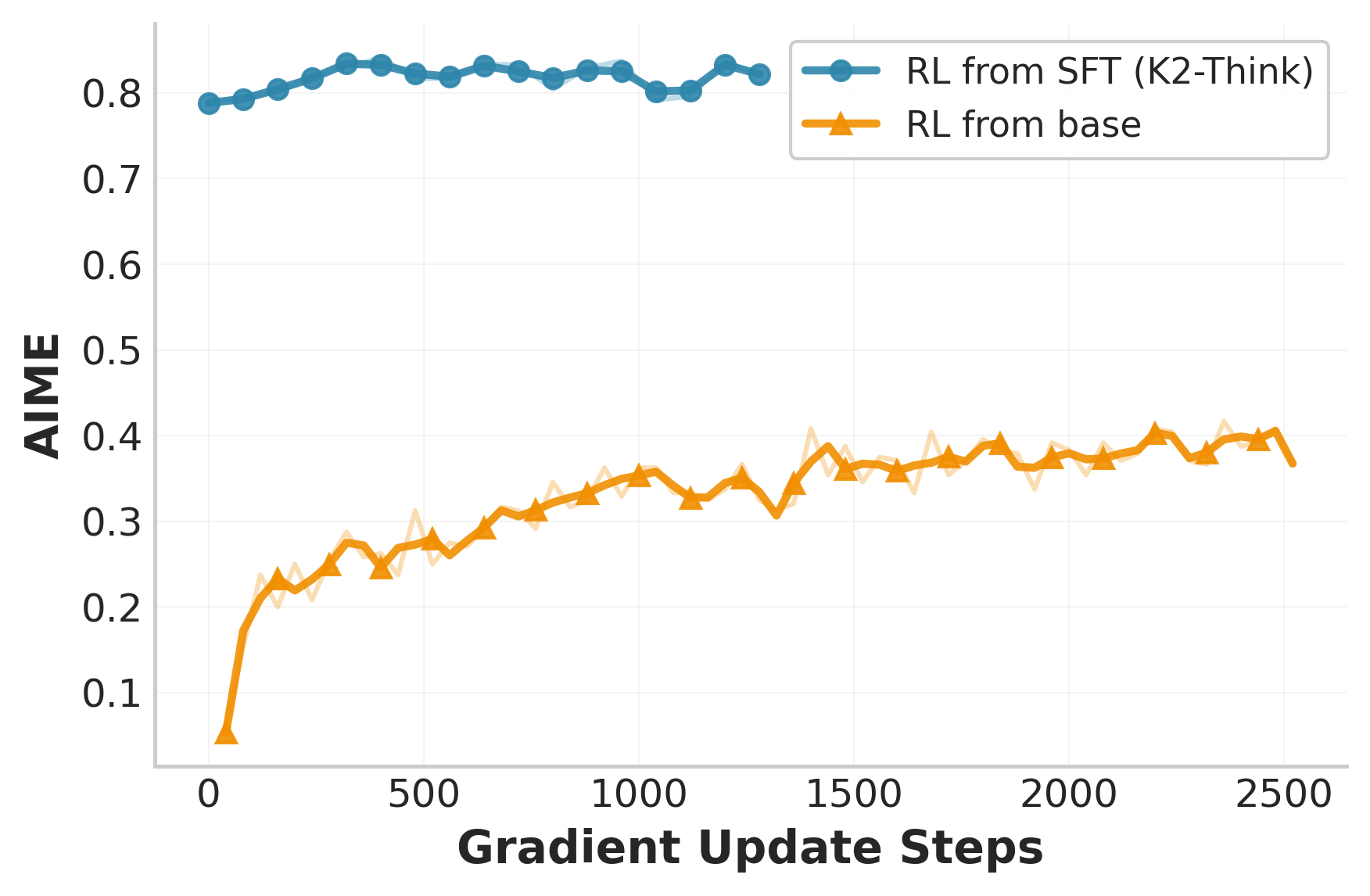}
        \label{fig:aime_base}
    \end{subfigure}
    \hfill
    \begin{subfigure}[b]{0.47\textwidth} 
        \centering
        \includegraphics[width=0.95\textwidth]{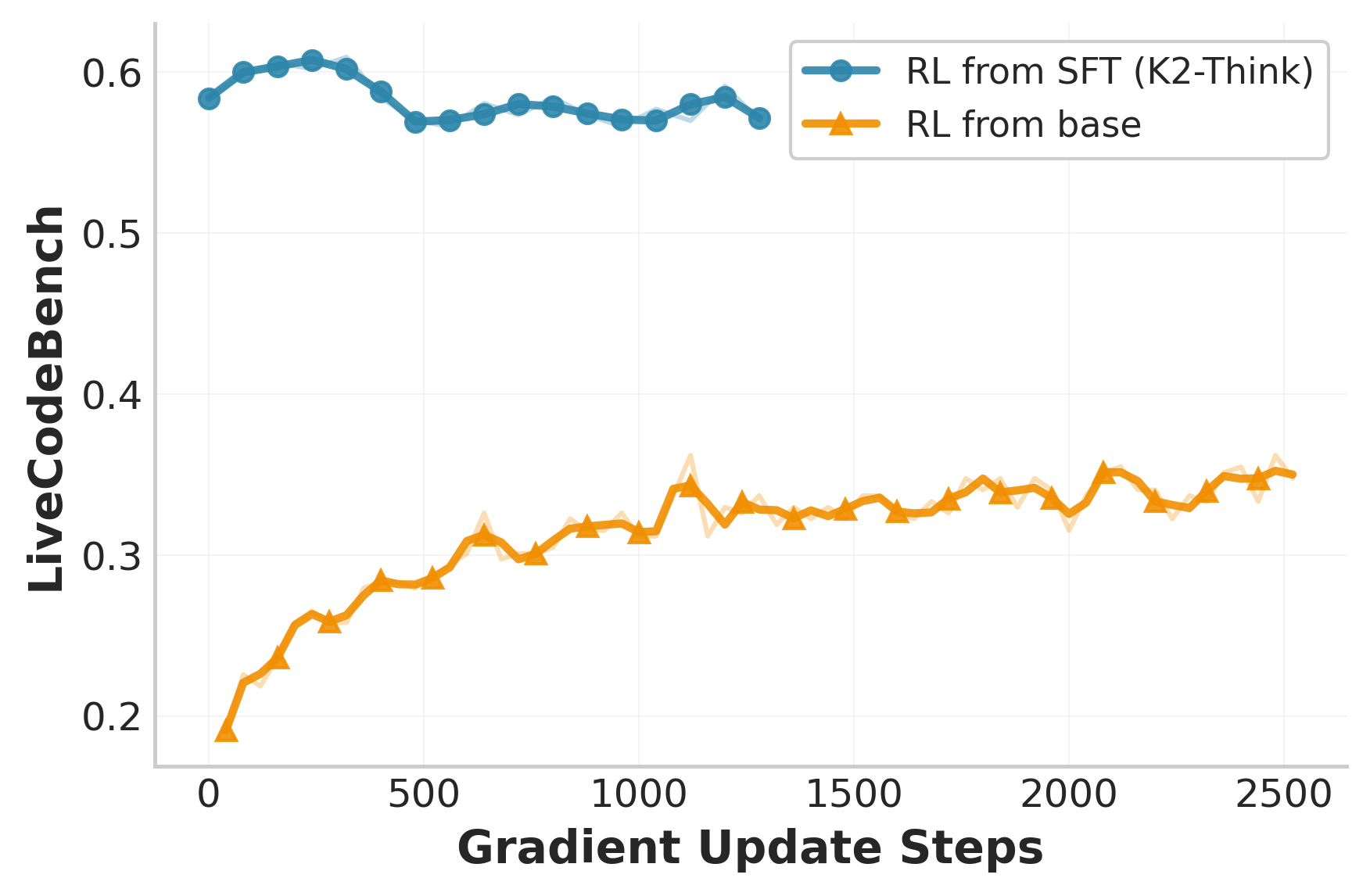}
        \label{fig:lcb_base}
    \end{subfigure}
    \vspace{0.8em} 
    \begin{subfigure}[b]{0.47\textwidth} 
        \centering
        \includegraphics[width=0.95\textwidth]{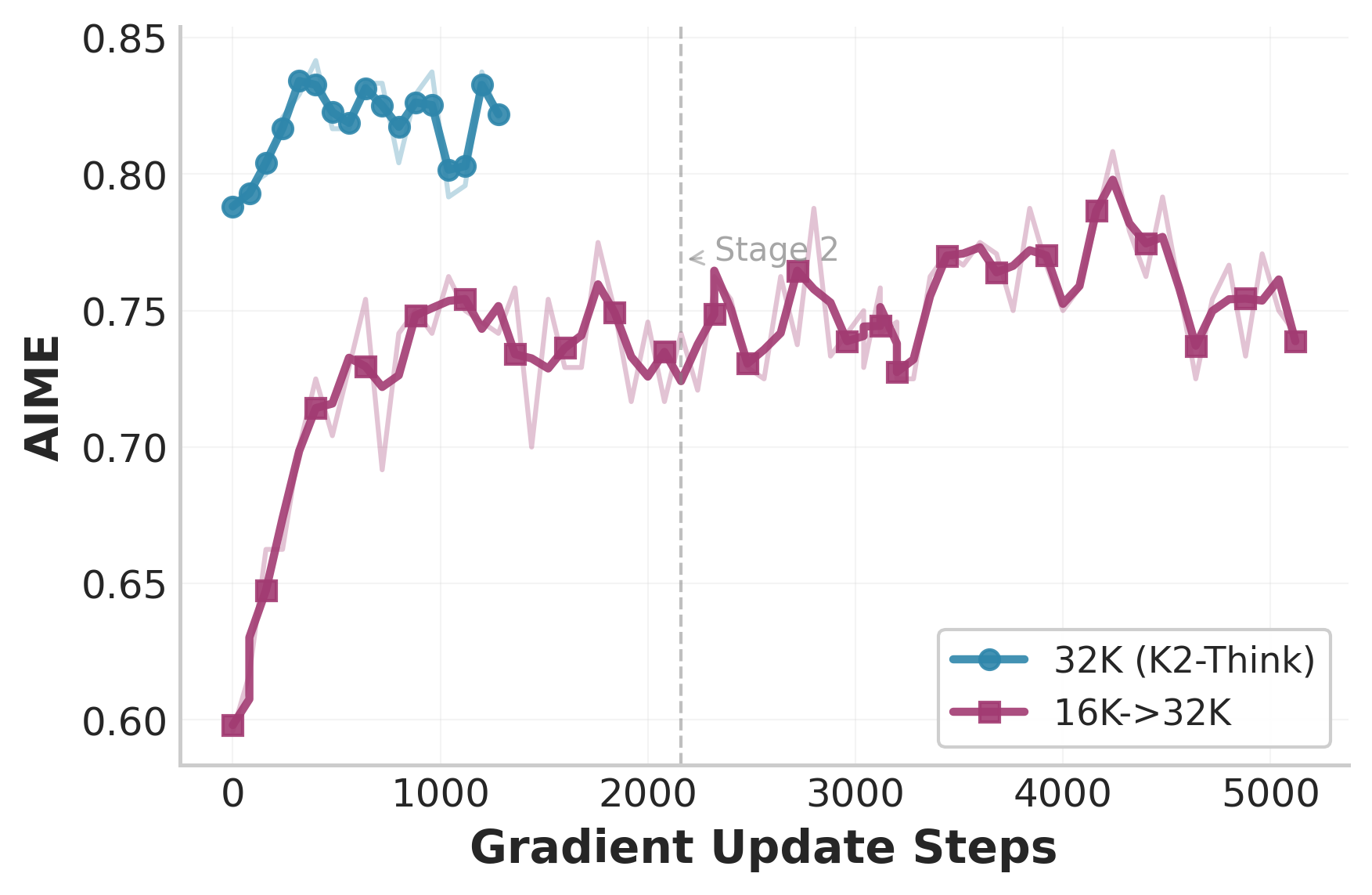}
        \label{fig:aime_multistage}
    \end{subfigure}
    \hfill 
    \begin{subfigure}[b]{0.47\textwidth} 
        \centering
        \includegraphics[width=0.95\textwidth]{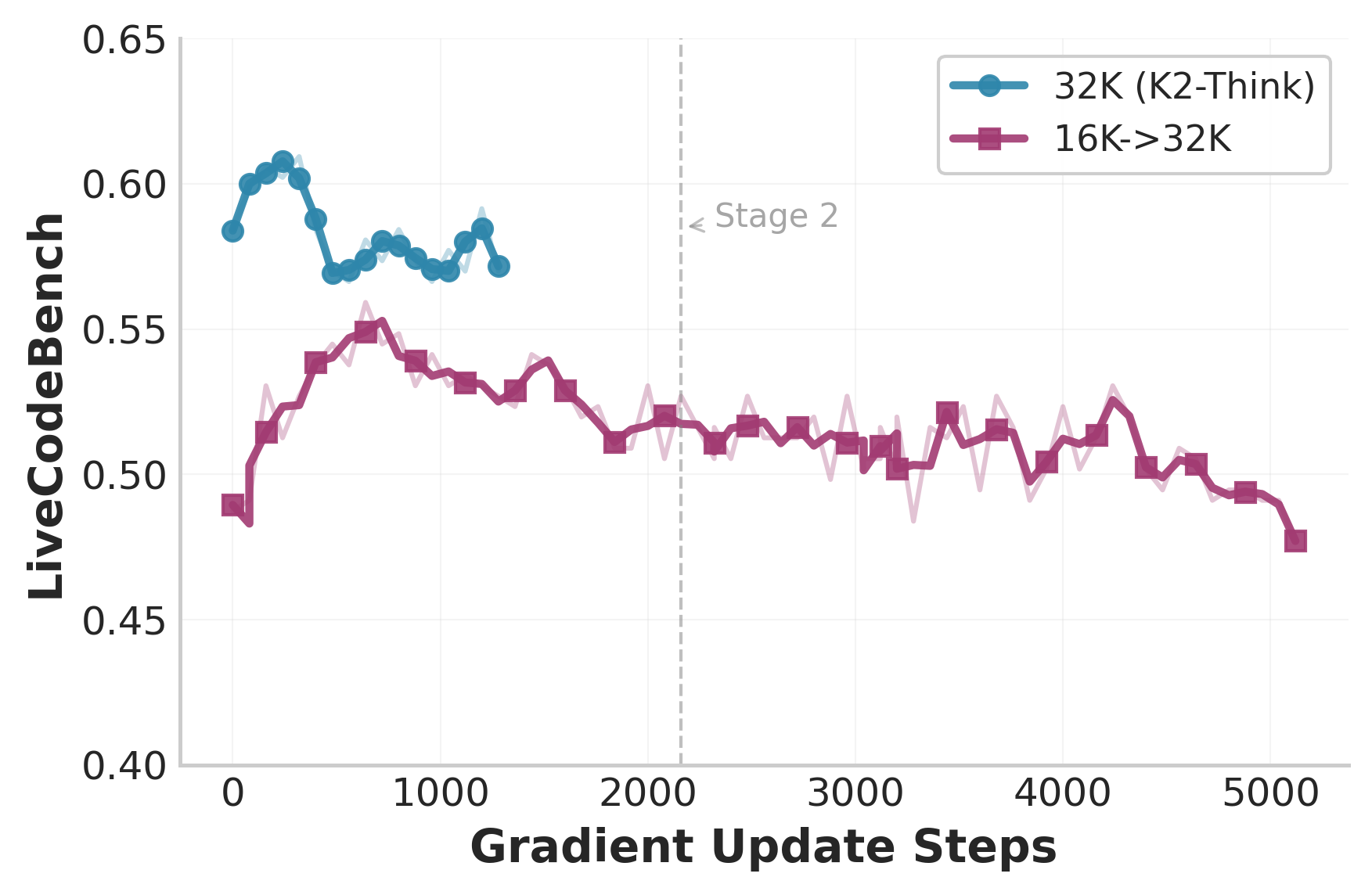}
        \label{fig:lcb_multistage}
    \end{subfigure}
    \caption{
        \textbf{Ablation Studies on Multi-stage Training and RL from Base Models.}
        \textbf{(top)}: RL from base models achieves much faster performance gains compared to RL from SFT models. However, a substantial performance gap remains, suggesting that SFT enhances the model's score at the cost of slower subsequent improvement and increased susceptibility to collapse during RL.
        \textbf{(bottom)}: reducing \ours{}-SFT maximum response length significantly impacts performance. Multi-stage training (16,000 to 32,000) struggles to recover original performance, even with prolonged training.
    }
    \label{fig:rl_main_plots}
\end{figure*}

For \ours{}’s RLVR, we use the Guru dataset~\citep{cheng2025revisiting}, which was curated to extend open-source reasoning models to verifiable domains beyond Math and Code. We leverage all six domains from the Guru dataset, comprising nearly 92,000 verifiable prompts that cover Math, Code, Science, Logic, Simulation, and Tabular tasks. We refer interested readers to the Guru paper for the detailed dataset curation, including de-duplication, reward designs, and filtering. Our RLVR implementation is built on the \texttt{verl} library~\citep{sheng2025hybridflow} with the GRPO algorithm~\citep{shao2024deepseekmath}.

\subsubsection{Observations}\label{sec:rl_wdw}
In this subsection we provide a retrospective set of observations that serve as motivation for future development. 

\paragraph{Starting from a strong SFT checkpoint yields better performance but limits RL gains.}

While RL consistently improves \ours{}-SFT performance across internal evaluations and public benchmarks, the absolute improvements were modest. As a comparative experiment, we also train a model with the same RL recipe and Guru data directly from the Qwen2.5-32B base. Figure~\ref{fig:rl_main_plots}~(top) demonstrates that RL training from the base model achieves nearly 40\% improvement on AIME 2024 over the training course, while RL from \ours{}-SFT yields only 5\% improvement. This validates that stronger SFT checkpoints leave less room for RL refinement, consistent with findings from \cite{liu2025acereason} regarding the relationship between SFT scope and subsequent RL effectiveness. Also, we notice RL training from the SFT checkpoint exhibits early plateauing and even degradation. We suspect that heavily ``SFTed'' models become constrained in their ability to explore alternative reasoning strategies during RL training, limiting the policy's capacity for meaningful adaptation.

\paragraph{Multi-stage RL training with reduced initial context length degrades performance.}

Many concurrent research efforts employ multi-stage training as implicit curriculum learning~\citep{Polaris2025,prolongRL,rastogi2025magistral}, incrementally increasing context length. This accelerates early training while the model develops competency and then allows the model in later stages to handle more difficult questions with the extended context. We test this approach by first constraining model output to 16,000 tokens during initial RL training from \ours{}-SFT, then expanding to 32,000 tokens (this is the maximum length seen during the SFT stage) for continued training.
As shown in Figure~\ref{fig:rl_main_plots}~(bottom), this multi-stage approach failed to match even the baseline SFT model performance. Cutting the maximum length below the SFT training configuration yields substantially lower performance. This negative result undermines the original motivation for multi-stage training to achieve on-par or better performance with shorter responses to save inference tokens.
We suspect that reducing context length below the SFT training regime (32k → 16k → 32k) disrupts the model's established reasoning patterns as we do not perform any additional data filtering to correspond to this multi-stage training. However, we do not evaluate expanding beyond the SFT context length (e.g., 32k → 48k), as implemented in Polaris~\citep{Polaris2025}, which may still provide benefits. 

\subsection{Phase 3: Test-time Improvement}\label{sec:ttc}

To further enhance \ours{} performance, we develop a test-time scaffolding that implements existing methods as well as integrates an original approach to provide structured input to our post-trained reasoning model. This subsection details two specific aspects of this scaffolding: agentic planning before reasoning, namely ``Plan-Before-You-Think'', and test-time scaling using Best-of-N sampling. These two techniques are pillars three and four of the complete \ours{} system. 

\begin{figure}[h!]
    \centering
    \includegraphics[width=\textwidth]{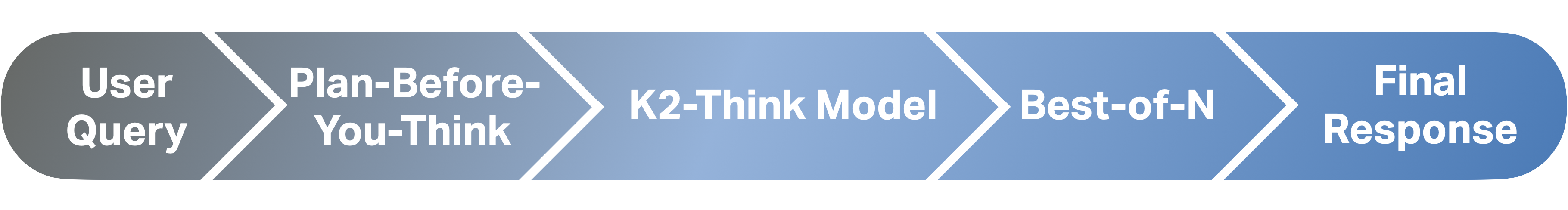}
    \caption{Schematic overview of how \ours{} generates responses via our test-time computation scaffold. A user query is first input to an external model which generates a high-level plan to provide a structured prompt to our \ours{} model. We then sample 3 responses, using an external model to select the best which is then provided as output.}
    \label{fig:k2_think_system}
\end{figure}

A diagram mapping the flow of information from the input provided, down to the final response, is illustrated in Figure~\ref{fig:k2_think_system}. First, the prompt is restructured to outline a high-level plan, highlighting relevant concepts. This augmented prompt is then passed through the \ours{} model, generating multiple responses. Finally, a pairwise comparison between candidate responses surfaces the best generation as the final output of our reasoning system. The remainder of this section provides details of how we set-up and implement each of these components.\footnote{Code for \ours{} test-time improvements is at: \rurl{https://github.com/MBZUAI-IFM/K2-Think-Inference}}

\paragraph{``Plan-Before-You-Think''.}

The first procedure of \ours{}'s test-time computation is the introduction of a planning agent. In our current system implementation, we simply ask the agent to extract key concepts from the query, and create a high-level plan from them. The generated plan is appended alongside the original query, and provided to the \ours{} model. \ours{}'s planning agent is simply implemented via prompting an instruction-tuned Language Model. We restrict this ``Plan-Before-You-Think'' procedure from providing direct answers or any reasoning trace. This deliberation phase, prior to any ``thinking'' by our reasoning model, has some basis in psychology and cognitive science. Planning and reasoning can be considered dual processes of human cognition and decision making~\citep{evans2010intuition} where planning is considered a meta-thinking process developing some structure to help guide one's thoughts. 

\paragraph{Best-of-N (BoN) sampling.}
Best-of-N sampling, sometimes called repeated sampling, is a method where an LLM generates N independent outputs for a given prompt, and a reward model (or verifier) chooses the best one according to some metric—such as accuracy, coherence, or alignment with human preference~\citep{stiennon2020learning,nakano2021webgpt}. This strategy effectively explores multiple possibilities and picks the most promising completion. 
 
Implementation-wise, we pick the answer by comparing the answer candidates pairwise, discarding the one that an independent LLM judges to be worse. In \ours{}, we finally adopt $N=3$, which provides a reasonable improvement with low cost. 

\subsubsection{Observations}\label{sec:ttc_wdw}

At inference time, we explore several approaches to enhance the \ours{} model's performance. We begin with simple engineering adjustments but soon discover that minor changes to our test-time computation procedures significantly impact overall performance.

We experiment with temperature tuning, iterating through a list of temperatures from 0.1 to 1.0, but find the overall improvement to be insignificant, leading us to use a temperature of 1.0 for all future runs. We also conduct extensive prompt engineering, trying over 30 different system prompts that utilized techniques like few-shot learning~\citep{brown2020language}, role-playing~\citep{kong2023better}, and situational prompting. However, we observe negligible gains. 

More sophisticated test-time scaling methods are also tried following~\cite{optillm}. We test several standard approaches including re2 (ReRead)~\citep{xu2024llavacot}, self-consistency~\citep{wang2023selfconsistency}, CoT with reflection~\citep{shinn2023reflexion}, and Mixture of Agents (MoA)~\citep{wang2024mixture}. Among these, Best-of-N (BoN) and MoA yield the most notable improvements. While MoA delivers marginally better performance, its significantly higher computational cost lead to the selection of BoN for the final \ours{} system. A different, more experimental approach involving Reinforcement Learning with rewards drawn from self-certainty signals~\citep{zhao2025learning} is also tested, but it does not lead to any improvement of the post-trained model's performance.

\subsection{Deploying K2-Think}\label{sec:serving}
We deploy \ours{} on Cerebras Wafer-Scale Engine (WSE) systems, leveraging the world's largest processor and speculative decoding~\citep{leviathan2023fast} to achieve unprecedented inference speeds for the reasoning system, making up the final two pillars of \ours{}. The WSE delivers approximately 2,000 tokens per second, representing a 10 times improvement over the nominal 200 tokens per second observed on typical deployment environments on a regular cloud provider. This dramatic speed-up fundamentally transforms the practical usability of long chain-of-thought reasoning.

Consider a typical complex reasoning task that generates a 32,000 token response, which is common for challenging mathematical proofs or multi-step coding problems. On the WSE, 32,000 token generation is completed in just 16 seconds, maintaining user engagement, enabling true back-and-forth problem solving, and providing a near real-time chat experience.

The performance advantage comes from the unique architecture of WSE. WSE keeps all model weights in massive on-chip memory, leveraging 25 Petabytes per second of on-chip memory bandwidth. Since auto-regressive models generate tokens serially, memory bandwidth can be a significant bottleneck during inference. By integrating greater compute, memory, and memory bandwidth in a single device, wafer-scale technology enables industry-leading inference speed for generative models.

This efficiency proves especially critical for our test-time computation approach and agent-based reasoning workflows. When performing best-of-3 sampling, the system must wait for all three responses to complete before LLM evaluation can select the optimal solution. 
Further, multi-step reasoning pipelines that require sequential calls for planning and generation suffer from cumulative delays. The WSE's low-latency inference keeps these workflows interactive, preventing the cascade of delays that would otherwise render complex reasoning tasks impractical.

The difference between waiting minutes versus seconds for each interaction fundamentally transforms the user experience from batch processing to interactive reasoning. This deployment ensures that \ours{} provides not just frontier reasoning capabilities but also the responsiveness required for practical, real-world applications, making sophisticated AI reasoning truly accessible for interactive use cases. We invite everyone to experience our \ours{} system, powered by Cerebras' WSE, via API and through \rurl{k2think.ai}.

\section{\ours{} Evaluation}\label{sec:evaluation}

We evaluate \ours{} in comparison with frontier models, both open-weight and proprietary, among a class of challenging reasoning benchmarks focused on Math, Code and Science. We design these evaluations to demonstrate that \ours{}, despite only having 32B parameters and fairly modest test-time computation, pushes the frontier of open-source reasoning models. In particular we find that \ours{} is highly capable for complex Math tasks, as shown in Table~\ref{tab:leaderboard}. In total we evaluate \ours{} on the following benchmarks:
\begin{itemize}
    \item \textsc{Math}
    \begin{itemize}
        \item \textbf{AIME 2024}~\citep{AIME2024}, \textbf{AIME 2025}~\citep{AIME2025}: The 2024 and 2025 editions of the American Invitation Mathematics Examination (AIME), with each year featuring 30 questions that have integer answers.
        \item \textbf{HMMT25}~\citep{balunovic2025matharena}: This dataset, used as part of the MathArena benchmarking suite, is drawn from the Harvard-MIT Mathematics Tournament February 2025 competition, featuring 30 questions drawn from the subject areas of Algebra+Number Theory, Combinatorics, and Geometry.
        \item \textbf{Omni-MATH-HARD} (Omni-HARD,~\cite{gao2024omni}): We use the most difficult subset of the Omni-MATH dataset, featuring questions sampled from competitive mathematics competitions at the Olympiad level from several countries, retaining only those problems that are rated as the top 2 difficulty levels (9.0 and 10.0). This set has 173 questions, a much larger competition math benchmark, and perhaps the most compelling one. 
        \item A global micro-average (Micro-Avg.) is obtained by dividing the total number of correct answers by the total number of questions across all datasets.
    \end{itemize}
    \item \textsc{Code}
    \begin{itemize}
        \item \textbf{LiveCodeBench} (LCBv5,~\cite{jain2024livecodebench}): A collection of 599 programming challenge problems aggregated from online platforms. We use queries aggregated between July 1, 2024 and February 1, 2025 (v5).
        \item \textbf{SciCode}~\citep{tian2024scicode}: SciCode evaluates a model's ability to generate code for solving 65 realistic scientific research questions, covering 16 subdomains from Physics, Math, Material Science, Biology, and Chemistry. We report scores from the version of the benchmark where background knowledge is included within the prompt. Since SciCode already performs a complex, multi-step planning phase in collating this information we do not run our ``Plan-Before-You-Think'' step during our evaluation of \ours{} on this baseline. All evaluation results include sub-problem and full-problem accuracies in Table~\ref{tab:leaderboard}.
    \end{itemize}
    \item \textsc{Science}
    \begin{itemize}
        \item \textbf{GPQA-Diamond} (GPQA-D,~\cite{gpqa}): This benchmark is comprised of 198 ``Google-proof'' advanced multiple-choice questions written by experts from biology, physics, and chemistry.
        \item \textbf{Humanity's Last Exam} (HLE,~\cite{phan2025humanity}): Humanity's Last Exam was developed by subject-matter experts and consists of 2158 multiple-choice and short-answer questions with solutions that are unambiguous and easily verifiable, but cannot be quickly answered via internet retrieval.
    \end{itemize}
\end{itemize}

We measure the performance of \ours{} in comparison to frontier reasoning models, both \emph{open-source} \{Qwen3-30B-A3B~\citep{yang2025qwen3}, GPT-OSS 20B~\citep{agarwal2025gpt}, QwQ-32B~\citep{team2025qwq}, OpenReasoning-Nemotron-32B~\citep{nvidia2025openreasoning}, DeepSeek R1~\citep{guo2025deepseekr1}, DeepSeek-v3.1 (Thinking)~\citep{deepseek_v31_release_2025}, GPT-OSS 120B~\citep{agarwal2025gpt}, Qwen3-235B-A22B (Thinking)~\citep{yang2025qwen3}\} and \emph{proprietary} \{GPT-5 (High)~\citep{openai_gpt5_2025}, Gemini-2.5 (Pro)~\citep{GoogleDeepMind2025Gemini}, o3 (High)~\citep{openaio3}\} to adequately assess the advancements made by our post-training and test-time computation scaffold. We use a standardized evaluation methodology across all benchmarks and models. The maximum generation length is set to 64,000 tokens, sampling temperature is fixed at 1.0, top-p is 0.95 and the stop token is \texttt{</answer>}. Each benchmark result reported in Table~\ref{tab:leaderboard} is the average of 16 independent pass@1 evaluations. 

\begin{table}[!htbp]
  \centering
\resizebox{0.98\textwidth}{!}{
\begin{NiceTabular}{l|ccccc|cc|cc}
\toprule
\textbf{\makecell{Benchmarks~$\rightarrow$}}
  & \multicolumn{5}{c|}{\textbf{Math}} 
  & \multicolumn{2}{c|}{\textbf{Code}} 
  & \multicolumn{2}{c}{\textbf{Science}} \\
\cmidrule(lr){2-6} \cmidrule(lr){7-8} \cmidrule(lr){9-10}
  \makecell{\textbf{Models}~$\downarrow$}& \textbf{AIME 2024} 
  & \textbf{AIME 2025} 
  & \textbf{HMMT25} 
  & \textbf{Omni-HARD} 
  & \textbf{Micro-Avg.}
  & \textbf{LCBv5} 
  & \textbf{SciCode (sub/main)}
  & \textbf{GPQA-D}
  & \textbf{HLE}\\
\midrule
\rowcolor[rgb]{ .863,  .937,  1}\textbf{\ours{}} 
  & 90.83 & 81.24 & 73.75 & 60.73 & 67.99 & 63.97 & 39.2 / 12.0 & 71.08 & 9.95 \\
\textbf{GPT-OSS 20B} 
  & 76.88 & 74.58 & 69.38 & 41.51 & 52.50 & 73.22 & 37.9 / 9.0 & 65.45 & 11.23 \\
\textbf{Qwen3-30B-A3B} 
  & 70.63 & 58.14 & 23.54 & 23.87 & 33.08 & 42.20 & 28.5 / 4.8 & 58.91 & 6.14 \\
\textbf{Nemotron 32B} 
  & 87.09 & 82.71 & 67.29 & 58.88 & 65.78 & 57.79 & 37.1 / 11.4 & 74.98 & 12.26 \\
\textbf{QwQ-32B} 
  & 79.38 & 69.17 & 51.46 & 46.93 & 53.69 & 65.22 & 36.9 / 11.5 & 66.24 & 9.98 \\
\textbf{GPT-OSS 120B} 
  & 89.58 & 84.59 & 81.88 & 57.76 & 67.20 & 74.53 & 38.8 / 11.0 & 77.04 & 18.58 \\
\textbf{Qwen3 235B-A22B} 
  & 86.68 & 75.43 & 61.88 & 56.91 & 62.99 & 56.64 & 39.3 / 10.9 & 65.55 & 14.23 \\
\textbf{DeepSeek V3.1}\textsuperscript{\textdagger}
  & 91.87 & 82.49 & 83.54 & 53.22 & 64.43 & 66.59 & 38.2 / 11.7 & 79.46 & 8.40 \\
\textbf{DeepSeek R1}\textsuperscript{\textdagger}
  & 74.38 & 65.21 & 47.08 & 51.33 & 55.06 & 61.01 & 36.7 / 11.5 & 71.08 & 8.50 * \\
  \midrule
\textbf{o3 High} 
  & 92.26 & 86.58 & 80.80 & 59.39 & 68.68 & 73.30 & 41.7 / 11.9 & 81.30 & 22.34 \\
\textbf{Gemini2.5 Pro} 
  & 87.24 & 85.75 & 74.18 & 69.36 & 73.82 & 58.24 & 45.1 / 15.4 & 84.51 & 19.93 \\
\textbf{GPT-5 High} 
  & 94.78 & 92.15 & 91.79 & 73.61 & 80.21 & 82.68 & 41.3 / 12.4 & 85.96 & 28.63 \\
\bottomrule
\end{NiceTabular}}
\vspace{-1.5mm}
\caption{Benchmark performance comparison of \ours{} against open-source (top) and proprietary (bottom) frontier models. All metrics are reported as percentages. \textbf{We find that \ours{} is especially strong on challenging Math benchmarks while also maintaining respectable performance on Code and Science.} Values marked with * are directly taken from published results. All other reported values are avg@16 accuracy of generated answers, evaluated locally or through paid API access. From these results, we see that our \ours{} system with only 32B parameters approaches or exceeds the performance of the frontier models that are orders of magnitude larger. \textit{\textdagger{} showing results for the original DeepSeek R1 and V3.1. Since the performance of DeepSeek R1-0528 is similar to V3.1, we do not report it separately.}
}
\label{tab:leaderboard}
\end{table}

\paragraph{\ours{} excels in competition math questions.}The evaluation results are summarized in Table~\ref{tab:leaderboard}. \ours{}, a 32B-parameter model, exhibits a micro-average score of \textbf{67.99} across all math questions. This result is particularly noteworthy when compared to other models of similar or slightly larger size, such as GPT-OSS 20B (m-avg. 52.50), Qwen3-30B-A3B (m-avg. 33.08), and OpenReasoning-Nemotron-32B (m-avg. 65.78). The results clearly show that \textbf{\ours{} surpasses these models by a significant margin}. Furthermore, \ours{}'s performance is not only dominant within its size class but also highly competitive with models that are orders of magnitude larger. Its math score also surpasses the larger models, including the two state-of-the-art open source models: DeepSeek V3.1 671B (m-avg. 64.43) and GPT-OSS 120B (m-avg. 67.20). Notably, \ours{} performs well on Omni-MATH-HARD (60.73), which contains the most difficult questions across competitions. \textbf{Such performance places \ours{} at the top of all open source models on math reasoning}, and is close to strong proprietary models such as o3 High, showing that \ours{} excels in the most challenging questions.

\paragraph{\ours{} is versatile in Science and Coding domains.} The evaluation results also show that \ours{} demonstrates a robust and competitive capability in both coding and scientific domains, solidifying its position as a versatile model. On coding benchmarks, \ours{} achieves a score of \textbf{63.97} on LiveCodeBench, significantly outperforming its similarly sized peers, including GPT-OSS 20B (42.20) and Qwen3-30B-A3B (36.9). This performance also surpasses the larger Qwen3 235B-A22B (56.64). When compared to larger models, \textbf{\ours{} shows parity and even superiority in certain metrics}: it achieves 39.2 on the SciCode benchmark (sub-problems), making it a close second compared with Qwen3 235B-A22B (39.3). On scientific reasoning, our system's performance on the GPQA-Diamond benchmark is 71.08, superior to most open-source models except OpenReasoning-Nemotron-32B (74.98), and GPT-OSS 120B (77.04). While its HLE score of 9.95 is not the highest, it remains respectable and indicative of a broad knowledge base. This combination of strong performance across diverse domains argues that \ours{} is not merely a specialist but a versatile model capable of tackling a wide range of analytical and knowledge-intensive tasks with high efficacy.

Beyond the preliminary conclusions shared in this report, our team is continuing to perform additional analyses and comparisons between \ours{} and a more complete set of competing models and reasoning benchmarks. It is however clear that \ours{} presents an advancement in open-source reasoning systems. With a 32B parameter model, and a moderate amount of test-time compute, our system provides comparative performance to models significantly larger (see Figure~\ref{fig:math_v_params} for a visual depiction). This level of parameter efficiency, in terms of benchmark performance is a notable achievement, specifically among complex mathematics reasoning tasks.

\paragraph{Component Analysis of \ours{} Test-Time Computation}

In order to analyze the individual contribution of each element of the test-time computation procedure to the final performance of \ours{}, we conduct an analysis where we implement each procedure in isolation on top of the post-trained checkpoint. That is, after performing both SFT and RL, we apply only the prompt restructuring via high-level planning or best-of-3 re-sampling and verification during evaluation. To simplify the discussion, we present this component analysis only using the four Math benchmarks however the overall insights are consistent across all other benchmarks.

\begin{table}[!h]
  \centering
  \vspace{3pt}
    \resizebox{0.85\textwidth}{!}{
    \begin{NiceTabular}{lcccc}
    \toprule
          & \multicolumn{1}{l}{\textbf{AIME 2024}} & \multicolumn{1}{l}{\textbf{AIME 2025}} & \multicolumn{1}{l}{\textbf{HMMT25}} & \multicolumn{1}{l}{\textbf{Omni-MATH-HARD}}\\
    \midrule
    \textbf{SFT+RL Checkpoint} & 86.26  & 77.72  & 66.46  & 56.74  \\
    \textbf{+ Plan only} & 85.21  & 81.04  & 71.87  & 58.97\\
    \textbf{+ Bo3 only} & 90.77  & 81.22  & 71.16  & 59.47 \\
    \textbf{+ Plan + Bo3 (\ours{})} & 90.83  & 81.24  & 73.75  & 60.73 \\
    \bottomrule
    \end{NiceTabular}%
    }
  \caption{Component analysis of the test-time computation procedures used to improve from our post-training checkpoint in the development of our final \ours{} system. The greatest gains come from Best-of-3 sampling,  further improvement is seen after combining with high-level planning.}
  \label{tab:ttc_component_analysis}%
\end{table}%

The component analyses presented here are executed in the same fashion as the comparative baselines contained in Table~\ref{tab:leaderboard}. All results presented in Table~\ref{tab:ttc_component_analysis} are averaged over 16 independent runs with the same settings as presented previously. We see in this analysis that the majority of the improvement over the post-trained checkpoint is afforded via Best-of-N scaling, using only 3 sampled generations per prompt. In isolation, the performance benefit of re-structuring the input prompt with a high-level plan also contributes an improvement to performance but with lesser effect. However, in combination with Best-of-N scaling the overall test-time procedure offers significant gains, offering 4-6 percentage points of improvement across all benchmarks.

\paragraph{``Plan-Before-You-Think'' Reduces Response Lengths}

With the complete \ours{} system, we require a model to create a plan before thinking. While we have shown this procedure to positively affect reasoning performance, the expansion of the prompt might cause more tokens used when formulating an answer. However, we found the opposite to be true. We report the average number of tokens \ours{} generated in the final response before and after implementing test-time computation in Table~\ref{tab:token_analysis} for select benchmarks evaluated across Math, Code and Science domains, comparing the \ours{} post-training checkpoint and the full system. The inclusion of this plan achieves two benefits: response quality improves and there is a reduction in the number of tokens used, by up to nearly 12\% in comparison to the post-training checkpoint. Thus, by conducting planning before reasoning, \ours{} provides more concise answers.

We also compare the average number of tokens used among the best performing open-weight models. We see that the \ours{} responses are much shorter than Qwen3-235-A22B and are in a range similar to the responses from GPT-OSS 120B in mathematical reasoning. When comparing response lengths for the code and science domains,  \ours{} are still shorter than Qwen3-235-A22B, but longer than GPT-OSS 120B.

\begin{table}[!t]
  \centering
  \vspace{3pt}
    \resizebox{0.95\textwidth}{!}{
    \begin{NiceTabular}{lcccccccccccc}
        \toprule
          \multicolumn{1}{c}{\textbf{Model}} & \multicolumn{2}{c}{\textbf{AIME 2024}} & \multicolumn{2}{c}{\textbf{AIME 2025}} & \multicolumn{2}{c}{\textbf{HMMT25}} & \multicolumn{2}{c}{\textbf{Omni-HARD}} & \multicolumn{2}{c}{\textbf{LCBv5}} & \multicolumn{2}{c}{\textbf{GPQA-D}}\\
    \cmidrule(lr){1-13}
    \textbf{SFT+RL Checkpoint} & 21,482 &  & 25,262 &  & 29,136 & & 34,042 & & 13,589 & & 14,998 & \\
    \textbf{\ours{}} & 20,040 & \cellcolor[rgb]{ .613,  .729,  .945}(-6.72\%) & 24,266 & \cellcolor[rgb]{ .706,  .776,  .906}(-3.94\%) & 27,030 & \cellcolor[rgb]{ .613,  .729,  .945}(-7.23\%) & 30,050 & \cellcolor[rgb]{ .557,  .663,  .859}(-11.73\%) & 12,166 & \cellcolor[rgb]{ .613,  .729,  .945}(-10.53\%) & 14,680 & \cellcolor[rgb]{ .906,  .902,  .932}(-2.12\%) \\
    \arrayrulecolor{gray}\midrule[0.1pt]
    \textbf{Qwen3-235B-A22B} & 29,896 &  & 34,541 &  & 39,767 & & 45,701 & & 27,716 & & 20,007 & \\
    \textbf{GPT-OSS 120B} & 15,971 &  & 19,151 &  & 25,566 & & 35,021 & & 7,389 & & 11,281 & \\
    \textbf{DeepSeek v3.1} & 12,364 &  & 15,143 &  & 19,073 & & 24,841 & & 6,158 & & 6,592 & \\
    \bottomrule
    \end{NiceTabular}%
    }
    \caption{An analytical comparison of the average number of tokens used between the full \ours{} system and the post-training checkpoint. After implementing our test-time computation scaffold, our response length decreases on average, with the percentage of reduction included in the shaded cells. We also compare to the average number of tokens generated by top-performing open-weight models, showing better efficiency than Qwen3-235B-A22B and similar to GPT-OSS 120B.}
  \label{tab:token_analysis}%
\end{table}%

\subsection{Red-teaming \ours{}}

Ensuring the safe operation of a model is essential for its open release. To this end, we systematically evaluate \ours{} against adversarial prompts, harmful content, and robustness stress tests using established public safety benchmarks \citep{lin2024achillesheelsurveyred}. For each benchmark, we sample 100 test cases and report a \textit{safe score}, where higher values indicate stronger safety performance. To provide a clear picture of real-world risks, we consolidate results into four key aspects that capture the practical safety surfaces most relevant in deployment:

\begin{enumerate}
    \item \textbf{High-Risk Content Refusal} --- ability to reject direct requests for unsafe or harmful outputs.
    \item \textbf{Conversational Robustness} --- maintaining safe behavior consistently across multi-turn dialogues.
    \item \textbf{Cybersecurity \& Data Protection} --- resilience against information leakage, prompt extraction, and cyberattack assistance.
    \item \textbf{Jailbreak Resistance} --- robustness to adversarial attacks designed to bypass safeguards.
\end{enumerate}

This framework provides a clearer operational safety profile and guides targeted mitigations.

\paragraph{High-Risk Content Refusal}

\begin{wraptable}{r}{0.45\textwidth}
\centering
\vspace{-\baselineskip} 
\begin{tabular}{@{}lc@{}}
\toprule
\textbf{Dataset} & \textbf{Score} \\
\midrule
Do-Not-Answer & 0.88 \\
HarmBench & 0.56 \\
PhysicalSafety & 0.49 \\
SimpleSafetyTests & 0.95 \\
ToxiGen & 0.97 \\
CoNA & 0.97 \\
HarmfulQ & 0.99 \\
\midrule
\textbf{Macro-average} & \textbf{0.83} \\
\bottomrule
\end{tabular}
\caption{High-risk content refusal results across safety datasets. The model achieves near-perfect performance on four of seven tasks, with clear improvement opportunities on HarmBench and PhysicalSafety.}
\label{tab:high_risk_refusal}
\vspace{-0.5cm}
\end{wraptable}

We first check the model's reliability in rejecting unsafe requests. Evaluation spans complementary datasets covering harmful instructions (\textbf{Do-Not-Answer} \citep{wang2023donotanswer}, \textbf{HarmBench} \citep{mazeika2024harmbench}), physical harm scenarios (\textbf{PhysicalSafety} \citep{2023Federicoarxiv:2309.07875v2}), basic safety checks (\textbf{SimpleSafetyTests} \citep{vidgen2023simplesafetytests}), toxic content generation (\textbf{ToxiGen} \citep{hartvigsen2022toxigen,hosseini2023empirical}), commonsense safety (\textbf{CoNA} \citep{2023Federicoarxiv:2309.07875v2}), and harmful Q\&A (\textbf{HarmfulQ} \citep{shaikh2023second}).

The results of this analysis are featured in Table~\ref{tab:high_risk_refusal}. \ours{} demonstrates extensive ability to avoid generating high-risk content as measured by near-perfect scores in 4 out of 7 benchmarks. Of the remaining 3 benchmarks in this aspect of safety evaluation, HarmBench and PhysicalSafety reveal a weakness in our system toward recognizing cyber or physical risks. We are actively working to improve our system along these dimensions of risk in its public facing deployment.

\paragraph{Conversational Robustness}
Next, we assess refusal consistency across multi-turn adversarial dialogues using \textbf{DialogueSafety} \citep{dinan-etal-2019-build}, \textbf{HH-RLHF} \citep{bai2022training}, and \textbf{DICES350} \citep{2023Loraarxiv:2306.11247v1} for dynamic dialogue manipulations.

We see in Table~\ref{tab:conversational_robustness} that \ours{} is especially robust to sustained adversarial dialogues and repeated efforts to ellicit harmful behaviors from our reasoning system. Here, \ours{} is near perfect at maintaining refusal consistency on both the DialogueSafety and HH-RLHF benchmarks.

\begin{table}[htbp]
\centering
\begin{minipage}{0.48\textwidth}
\centering
\begin{tabular}{@{}lc@{}}
\toprule
\textbf{Dataset} & \textbf{Score} \\
\midrule
DialogueSafety & 0.99 \\
HH-RLHF & 0.95 \\
DICES350 & 0.73 \\
\midrule
\textbf{Macro-average} & \textbf{0.89} \\
\bottomrule
\end{tabular}
\caption{Conversational robustness results across dialogue safety datasets. The model exhibits notable robustness to multi-turn adversarial attempts to produce harmful outputs, with particular strength on DialogueSafety and room for improvement on DICES350.}
\label{tab:conversational_robustness}
\end{minipage}
\hfill
\begin{minipage}{0.48\textwidth}
\centering
\begin{tabular}{@{}lc@{}}
\toprule
\textbf{Dataset} & \textbf{Score} \\
\midrule
PersonalInfoLeak (few-shot) & 0.86 \\
CyberattackAssistance & 0.47 \\
PromptExtractionRobustness & 0.35 \\
\midrule
\textbf{Macro-average} & \textbf{0.56} \\
\bottomrule
\end{tabular}
\caption{Cybersecurity, data protection, and prompt extraction results. The model demonstrates robustness against leaking personal information, with significant room for improvement on cyberattack assistance prevention and prompt extraction robustness.}
\label{tab:cybersecurity}
\end{minipage}
\end{table}

\paragraph{Cybersecurity \& Data Protection \& Prompt Extraction}
We evaluate resilience against data leakage and misuse with \textbf{PersonalInfoLeak} \citep{2023Haoranarxiv:2304.05197v3} (privacy leakage), \textbf{CyberattackAssistance} \citep{bhatt2023purplellamacybersecevalsecure} (hacking assistance), and \textbf{PromptExtractionRobustness} \citep{2023Samarxiv:2311.01011v1} (system prompt extraction).

We see in Table~\ref{tab:cybersecurity} that \ours{} is able to resist attempts to extract personally identifying information while unfortunately exhibiting some susceptibility to revealing the system prompt and aiding in devising cyberattacks. This indicates an opportunity to further tune our reasoning system for improved resilience.

\paragraph{Jailbreak Resistance}

\begin{wraptable}{r}{0.5\textwidth}
\centering
\vspace{-1.15\baselineskip} 
\begin{tabular}{@{}lc@{}}
\toprule
\textbf{Dataset} & \textbf{Score} \\
\midrule
Few-Shot Attack & 0.96 \\
Gandalf Ignore & 0.87 \\
Tense Change & 0.84 \\
Multilingual & 0.83 \\
PromptInjection & 0.77 \\
One-Sided Statement & 0.77 \\
Refusal Suppression & 0.76 \\
Persona Modulation & 0.59 \\
Do-Anything-Now & 0.43 \\
LatentJailbreak & 0.37 \\
\midrule
\textbf{Macro-average} & \textbf{0.72} \\
\bottomrule
\end{tabular}
\caption{Jailbreak resistance results across adversarial prompt techniques. The model demonstrates mixed resilience, with strong performance against direct attacks and vulnerabilities to indirect methods.}
\label{tab:jailbreak_resilience}
\vspace{-0.5cm}
\end{wraptable}

Finally, we evaluate various adversarial attack strategies: hidden triggers (\textbf{LatentJailbreak} \citep{2023Huachuanarxiv:2307.08487v3}), prompt redirection (\textbf{PromptInjection} \citep{liu2023promptinjection}), instruction overrides (\textbf{Gandalf Ignore} \citep{schulhoff-etal-2023-ignore}), role-play attacks (\textbf{DAN} \citep{shen2023do}), cross-lingual exploits (\textbf{Multilingual} \citep{2023Wenxuanarxiv:2310.00905v1}), grammatical perturbations (\textbf{Tense Change} \cite{lin2024achillesheelsurveyred}), adversarial demonstrations (\textbf{Few-Shot Attack} \citep{2023Zemingarxiv:2310.06387v1}), bias-driven attacks (\textbf{One-Sided Statement} \citep{liu2023Goal-Oriented}), identity manipulation (\textbf{Persona Modulation} \citep{Shah2023personamodulation}), and direct refusal bypasses (\textbf{Refusal Suppression} \citep{wei2023jailbroken}).

\ours's jailbreak resistance results (shown in Table~\ref{tab:jailbreak_resilience}) demonstrate a mixture of resilience and susceptibility to various adversarial prompt strategies. \ours{} exhibits strong performance when attacks are immediately apparent but shows an apparent weakness to indirect attacks. This lack of generalized robustness to adversarial jailbreaking attempts illustrates a need to thoroughly improve our publicly deployed reasoning system.

\paragraph{Overall Results}
Across all four dimensions, results are aggregated into a single Safety-4 macro score, computing the average from the four analyses performed as part of our safety testing of \ours{}. The macro average of each of the four analyses are included in Table~\ref{tab:safety4macro}.

\begin{table}[h]
\centering
\begin{tabular}{l r}
\toprule
\textbf{Safety Aspect} & \textbf{Macro-Avg Score} \\
\midrule
High-Risk Content Refusal & 0.83 \\
Conversational Robustness & 0.89 \\
Cybersecurity \& Data Protection & 0.56 \\
Jailbreak Resistance & 0.72 \\
\midrule
\textbf{Safety-4 Macro (avg)} & \textbf{0.75} \\
\bottomrule
\end{tabular}
\caption{Overall Safety-4 results which is a composite score of the four safety surfaces evaluated in this broad analysis. The macro score of 0.75 indicates that \ours{} establishes a solid safety profile with specific strengths in harmful content refusal and maintaining consistent behavior in conversations.}
\label{tab:safety4macro}
\end{table}

Overall, \ours{} achieves a \textbf{Safety-4 macro score of 0.75}, indicating a solid baseline of safety with strong performance in refusing harmful content and maintaining consistent behavior in conversations. At the same time, we recognize that further work is required to strengthen \textbf{cybersecurity defenses, jailbreak robustness, and refusal calibration}. While establishing a solid baseline, we acknowledge clear opportunities to improve the safety of our reasoning system. Addressing these areas is an active priority in our roadmap to further improve \ours{} under adversarial conditions.

\section{Related Work}\label{sec:related_work}
\paragraph{Extending base language model capabilities via SFT}

Supervised fine-tuning (SFT) has become a widely used post-training method to extend the capability boundary of Large Language Models~\citep{InstructGPT, llama3, guo2025deepseekr1, llama-nemotron}. Early SFT work primarily focused on task specialization, adapting foundational models to specific NLP benchmarks like text classification or translation on narrowly-defined datasets~\citep{roberta, raffel2020exploring}. This paradigm shifted significantly with the rise of large-scale instruction tuning; the goal evolved from single-task mastery to creating general-purpose assistants capable of following diverse human commands~\citep{flan, InstructGPT, alpaca}. More recently, SFT has pivoted towards enhancing complex reasoning on diverse downstream tasks like math, code, and science~\citep{qwen2.5code, qwen2.5math, phi4reasoning, liuliftreasoning}. Some approaches focus on scale, constructing massive datasets of reasoning traces to instill robust, long-chain-of-thought capabilities in models~\citep{guha2025openthoughts, tian2025not, liu2025acereason}. In contrast, other methods demonstrate that meticulously curated, high-quality data can also endow LLMs with expert-level reasoning in domains like math~\citep{limo, s1}. Building on the above, our work conducts analysis and provides practical insights on SFT. 

\paragraph{Improving LLM Reasoning with RL}
Reinforcement Learning from Verifiable Rewards~(RLVR) has emerged as a powerful paradigm for enhancing the reasoning capabilities of Large Language Models~\citep{guo2025deepseekr1,OpenAIo1}. Following initial successes, a significant body of open work has explored RLVR, primarily concentrating on specializing models for highly challenging single domains. Efforts such as Open-Reasoner-Zero~\citep{hu2025open}, Skywork-OR1~\citep{skywork-or1-2025}, DeepScaler~\citep{deepscaler2025}, and SimpleRL~\citep{zeng2025simplerl} have notably leveraged extensive mathematical data to achieve state-of-the-art performance on complex math benchmarks. Similarly, DeepCoder~\citep{deepcoder2025} focused on RL for code generation tasks. While powerful within their specific areas, this domain-specific focus inherently limits the generalizability of the resulting models across the broader landscape of reasoning tasks. Concurrent works to our \ours{} development like General-Reasoner~\citep{generalreasoner} and Nemotron-CrossThinker~\citep{akter2025nemotron} have begun to explore broader domains for RL training. However, none of these works explore the added utility of test-time computation for improving the general reasoning capabilties of post-trained models.

\paragraph{Test Time Scaling} Test-time scaling has been a major component of proprietary models released in recent years; such as o1~\citep{OpenAIo1}, Grok Heavy~\citep{xAI2025Grok4}, Gemini 2.5~\citep{GoogleDeepMind2025Gemini}, and GPT-5~\citep{openai_gpt5_2025}. However, with fairly little transparency about specific components and their overall effect. The closest work to ours is PlanGEN~\citep{parmar2025plangen}, a multi-model framework for planning and reasoning combining a constraint model, a verification model, and a selection model to guide inference-time algorithms including Best-of-N. By using constraint-guided iterative verification and a modified UCB-based selection policy, PlanGEN chooses the most suitable algorithm for each problem instance. Importantly, they use Best-of-N with verifiers on the \textit{plans}: we use it for the generated \textit{responses}.

Also related are general LLM-based hierarchical reasoning approaches, particularly those that operate with at least one level of hierarchy doing planning. \cite{wang2024coplanner} has a planning model provide high-level strategy while a solver model performs detailed reasoning. HyperTree Planning~\citep{gui2025hypertree} models planning with a hypertree-structure, allowing LLMs to decompose planning queries into structured sub-tasks. \cite{wang2025hrm} demonstrates a brain-inspired architecture with separate recurrent modules for high-level planning and low-level reasoning, showing that explicit separation of timescales improves performance on algorithmic reasoning tasks. Our novelty is to combine our ``Plan-Before-You-Think'' approach, a type of multi-LLM-hierarchical reasoning, with Best-of-N with verifiers~\citep{cobbe2021training} in order to return the best responses.

\section{Discussion}
\label{sec:discussion}

\subsection{Primary technical insights}

\paragraph{Multiple domains are important for post-training.} Following the findings from Guru~\citep{cheng2025revisiting}, there is a need to expand post-training to include more domains for general reasoning models. The effect of post-training, and the domains utilized, is nuanced. Domains commonly included in pre-training (Math, Code, and Science) broadly benefit from a variety of post-training data as the refinement of the model's chains of thought is supported by the knowledge it already has. However those domains with limited pre-training exposure--like Logic and Simulation tasks--only improve when they are included in the RL training pipeline. This indicates that using diverse, multi-domain datasets is critical for developing truly versatile reasoning models.

\paragraph{Test-time computation performance gains can be additive with the right combination.} We find that two simple test-time computation procedures work well together: our ``Plan-Before-You-Think'' prompt restructuring in conjunction with Best-of-N scaling. Each individual method does improve over the \ours{} model but the largest gains in performance are seen when these components are combined. To our surprise, simply extracting a high level plan focused on the core concepts associated with the input and only sampling 3 candidate responses are sufficient to provide significant improvement.

\paragraph{``Plan-Before-You-Think'' improves model performance while reducing token expenditure.}
By requiring the model to create a plan before initiating its reasoning process, we achieve two benefits: planning itself improves response quality, and response lengths are reduced by nearly 12\%.  

\subsection{Looking forward}

\paragraph{Empowering small models to ``punch above their weight''.} With the complete \ours{} system, we demonstrate that a 32B-scale model, post-trained to produce long reasoning chains of thought, paired with relatively little test-time computation can endow the small model with capabilities that are competitive with models with orders of magnitude more parameters. Altogether our end-to-end reasoning system unlocks performance at the frontier of current open-source capabilities.

\paragraph{Beyond Open Source.} We are extending the limit of our open-source activities beyond data, models and training artifacts. This expansion of our open-source efforts will now include deploying our full reasoning system for public use. We are publishing our test-time computation implementation as well. \ours{} is broadly available via API and an online web portal. In this we are opening avenues to explore how to best ``battle-test'' public facing LLM infrastructure. Details about how to use and interact with \ours{} can be found at \rurl{k2think.ai}, we proudly invite all to try it out!

\ours{} is a compelling stepping stone for our ongoing efforts to broaden access to foundation model research and development through open-science. Our motivation to deploy \ours{} for public use is grounded in curiosity about how to best engineer inference systems for large-scale foundation models. 
Secondarily, as we continue scaling our own open-source models, there will be a time when simply making the weights and training artifacts public is no longer useful as fewer institutions and organizations will be able to host or interact with the models themselves. This by-product of our work, investigating and building ever more capable open models, is antithetical to our founding ethos as a research institute. We are committed to making publicly available as much of our model development and deployment as possible in order to enable all who are interested to build on or contribute to our work. The lessons we learn through deployment with \ours{} will be critical to our ongoing development of larger and more capable models.

\section*{Acknowledgment}

The authors hereby acknowledge and thank the strong support and collaboration of G42 for their contributions throughout the project, including the essential computational infrastructure as well as significant expertise in evaluation methodology and safety protocols. This partnership proved instrumental in advancing our research objectives.

\newpage
\bibliographystyle{plainnat}
\bibliography{references}

\begin{thebibliography}{102}
\providecommand{\natexlab}[1]{#1}
\providecommand{\url}[1]{\texttt{#1}}
\expandafter\ifx\csname urlstyle\endcsname\relax
  \providecommand{\doi}[1]{doi: #1}\else
  \providecommand{\doi}{doi: \begingroup \urlstyle{rm}\Url}\fi

\bibitem[Abdin et~al.(2025)Abdin, Agarwal, Awadallah, Balachandran, Behl, Chen, de~Rosa, Gunasekar, Javaheripi, Joshi, et~al.]{phi4reasoning}
Marah Abdin, Sahaj Agarwal, Ahmed Awadallah, Vidhisha Balachandran, Harkirat Behl, Lingjiao Chen, Gustavo de~Rosa, Suriya Gunasekar, Mojan Javaheripi, Neel Joshi, et~al.
\newblock Phi-4-reasoning technical report.
\newblock \emph{arXiv preprint arXiv:2504.21318}, 2025.

\bibitem[Agarwal et~al.(2025{\natexlab{a}})Agarwal, Ahmad, Ai, Altman, Applebaum, Arbus, Arora, Bai, Baker, Bao, et~al.]{agarwal2025gpt}
Sandhini Agarwal, Lama Ahmad, Jason Ai, Sam Altman, Andy Applebaum, Edwin Arbus, Rahul~K Arora, Yu~Bai, Bowen Baker, Haiming Bao, et~al.
\newblock gpt-oss-120b \& gpt-oss-20b model card.
\newblock \emph{arXiv preprint arXiv:2508.10925}, 2025{\natexlab{a}}.

\bibitem[Agarwal et~al.(2025{\natexlab{b}})Agarwal, Zhang, Yuan, Han, and Peng]{min-entropy}
Shivam Agarwal, Zimin Zhang, Lifan Yuan, Jiawei Han, and Hao Peng.
\newblock The unreasonable effectiveness of entropy minimization in llm reasoning, 2025{\natexlab{b}}.
\newblock URL \url{https://arxiv.org/abs/2505.15134}.

\bibitem[Akter et~al.(2025)Akter, Prabhumoye, Novikov, Han, Lin, Bakhturi, Nyberg, Choi, Patwary, Shoeybi, et~al.]{akter2025nemotron}
Syeda~Nahida Akter, Shrimai Prabhumoye, Matvei Novikov, Seungju Han, Ying Lin, Evelina Bakhturi, Eric Nyberg, Yejin Choi, Mostofa Patwary, Mohammad Shoeybi, et~al.
\newblock Nemotron-crossthink: Scaling self-learning beyond math reasoning.
\newblock \emph{arXiv preprint arXiv:2504.13941}, 2025.

\bibitem[An et~al.(2025)An, Xie, Li, Li, Zhang, Gong, Zhong, Xu, Qiu, Wang, and Kong]{Polaris2025}
Chenxin An, Zhihui Xie, Xiaonan Li, Lei Li, Jun Zhang, Shansan Gong, Ming Zhong, Jingjing Xu, Xipeng Qiu, Mingxuan Wang, and Lingpeng Kong.
\newblock Polaris: A post-training recipe for scaling reinforcement learning on advanced reasoning models, 2025.
\newblock URL \url{https://hkunlp.github.io/blog/2025/Polaris}.

\bibitem[Aroyo et~al.(2023)Aroyo, Taylor, D{\'{\i}}az, Homan, Parrish, Serapio{-}Garc{\'{\i}}a, Prabhakaran, and Wang]{2023Loraarxiv:2306.11247v1}
Lora Aroyo, Alex~S. Taylor, Mark D{\'{\i}}az, Christopher Homan, Alicia Parrish, Gregory Serapio{-}Garc{\'{\i}}a, Vinodkumar Prabhakaran, and Ding Wang.
\newblock {DICES} dataset: Diversity in conversational {AI} evaluation for safety.
\newblock In \emph{Advances in Neural Information Processing Systems 36: Annual Conference on Neural Information Processing Systems 2023, NeurIPS 2023, New Orleans, LA, USA, December 10 - 16, 2023}, 2023.
\newblock URL \url{http://papers.nips.cc/paper\\_files/paper/2023/hash/a74b697bce4cac6c91896372abaa8863-Abstract-Datasets\\_and\\_Benchmarks.html}.

\bibitem[Bai et~al.(2022)Bai, Jones, Ndousse, Askell, Chen, DasSarma, Drain, Fort, Ganguli, Henighan, et~al.]{bai2022training}
Yuntao Bai, Andy Jones, Kamal Ndousse, Amanda Askell, Anna Chen, Nova DasSarma, Dawn Drain, Stanislav Fort, Deep Ganguli, Tom Henighan, et~al.
\newblock Training a helpful and harmless assistant with reinforcement learning from human feedback, 2022.
\newblock URL \url{https://doi.org/10.48550/arXiv.2204.05862}.

\bibitem[Balunovi{\'c} et~al.(2025)Balunovi{\'c}, Dekoninck, Petrov, Jovanovi{\'c}, and Vechev]{balunovic2025matharena}
Mislav Balunovi{\'c}, Jasper Dekoninck, Ivo Petrov, Nikola Jovanovi{\'c}, and Martin Vechev.
\newblock Matharena: Evaluating llms on uncontaminated math competitions.
\newblock \emph{arXiv preprint arXiv:2505.23281}, 2025.

\bibitem[Bercovich et~al.(2025)Bercovich, Levy, Golan, Dabbah, El-Yaniv, Puny, Galil, Moshe, Ronen, Nabwani, Shahaf, Tropp, Karpas, Zilberstein, Zeng, Singhal, Bukharin, Zhang, Konuk, Shen, Mahabaleshwarkar, Kartal, Suhara, Delalleau, Chen, Wang, Mosallanezhad, Renduchintala, Qian, Rekesh, Jia, Majumdar, Noroozi, Ahmad, Narenthiran, Ficek, Samadi, Huang, Jain, Gitman, Moshkov, Du, Toshniwal, Armstrong, Kisacanin, Novikov, Gitman, Bakhturina, Scowcroft, Kamalu, Su, Kong, Kliegl, Karimi, Lin, Satheesh, Parmar, Gundecha, Norick, Jennings, Prabhumoye, Akter, Patwary, Khattar, Narayanan, Waleffe, Zhang, Su, Huang, Kong, Chadha, Jain, Harvey, Segal, Huang, Kashirsky, McQueen, Putterman, Lam, Venkatesan, Wu, Nguyen, Kilaru, Wang, Warno, Somasamudramath, Bhaskar, Dong, Assaf, Mor, Argov, Junkin, Romanenko, Larroy, Katariya, Rovinelli, Balas, Edelman, Bhiwandiwalla, Subramaniam, Ithape, Ramamoorthy, Wu, Velury, Almog, Daw, Fridman, Galinkin, Evans, Ghosh, Luna, Derczynski, Pope, Long, Schneider, Siman, Grzegorzek,
  Ribalta, Katariya, Alexiuk, Conway, Saar, Guan, Pawelec, Prayaga, Kuchaiev, Ginsburg, Olabiyi, Briski, Cohen, Catanzaro, Alben, Geifman, and Chung]{llama-nemotron}
Akhiad Bercovich, Itay Levy, Izik Golan, Mohammad Dabbah, Ran El-Yaniv, Omri Puny, Ido Galil, Zach Moshe, Tomer Ronen, Najeeb Nabwani, Ido Shahaf, Oren Tropp, Ehud Karpas, Ran Zilberstein, Jiaqi Zeng, Soumye Singhal, Alexander Bukharin, Yian Zhang, Tugrul Konuk, Gerald Shen, Ameya~Sunil Mahabaleshwarkar, Bilal Kartal, Yoshi Suhara, Olivier Delalleau, Zijia Chen, Zhilin Wang, David Mosallanezhad, Adi Renduchintala, Haifeng Qian, Dima Rekesh, Fei Jia, Somshubra Majumdar, Vahid Noroozi, Wasi~Uddin Ahmad, Sean Narenthiran, Aleksander Ficek, Mehrzad Samadi, Jocelyn Huang, Siddhartha Jain, Igor Gitman, Ivan Moshkov, Wei Du, Shubham Toshniwal, George Armstrong, Branislav Kisacanin, Matvei Novikov, Daria Gitman, Evelina Bakhturina, Jane~Polak Scowcroft, John Kamalu, Dan Su, Kezhi Kong, Markus Kliegl, Rabeeh Karimi, Ying Lin, Sanjeev Satheesh, Jupinder Parmar, Pritam Gundecha, Brandon Norick, Joseph Jennings, Shrimai Prabhumoye, Syeda~Nahida Akter, Mostofa Patwary, Abhinav Khattar, Deepak Narayanan, Roger Waleffe,
  Jimmy Zhang, Bor-Yiing Su, Guyue Huang, Terry Kong, Parth Chadha, Sahil Jain, Christine Harvey, Elad Segal, Jining Huang, Sergey Kashirsky, Robert McQueen, Izzy Putterman, George Lam, Arun Venkatesan, Sherry Wu, Vinh Nguyen, Manoj Kilaru, Andrew Wang, Anna Warno, Abhilash Somasamudramath, Sandip Bhaskar, Maka Dong, Nave Assaf, Shahar Mor, Omer~Ullman Argov, Scot Junkin, Oleksandr Romanenko, Pedro Larroy, Monika Katariya, Marco Rovinelli, Viji Balas, Nicholas Edelman, Anahita Bhiwandiwalla, Muthu Subramaniam, Smita Ithape, Karthik Ramamoorthy, Yuting Wu, Suguna~Varshini Velury, Omri Almog, Joyjit Daw, Denys Fridman, Erick Galinkin, Michael Evans, Shaona Ghosh, Katherine Luna, Leon Derczynski, Nikki Pope, Eileen Long, Seth Schneider, Guillermo Siman, Tomasz Grzegorzek, Pablo Ribalta, Monika Katariya, Chris Alexiuk, Joey Conway, Trisha Saar, Ann Guan, Krzysztof Pawelec, Shyamala Prayaga, Oleksii Kuchaiev, Boris Ginsburg, Oluwatobi Olabiyi, Kari Briski, Jonathan Cohen, Bryan Catanzaro, Jonah Alben, Yonatan
  Geifman, and Eric Chung.
\newblock Llama-nemotron: Efficient reasoning models, 2025.
\newblock URL \url{https://arxiv.org/abs/2505.00949}.

\bibitem[Bhatt et~al.(2023)Bhatt, Chennabasappa, Nikolaidis, Wan, Evtimov, Gabi, Song, Ahmad, Aschermann, Fontana, Frolov, Giri, Kapil, Kozyrakis, LeBlanc, Milazzo, Straumann, Synnaeve, Vontimitta, Whitman, and Saxe]{bhatt2023purplellamacybersecevalsecure}
Manish Bhatt, Sahana Chennabasappa, Cyrus Nikolaidis, Shengye Wan, Ivan Evtimov, Dominik Gabi, Daniel Song, Faizan Ahmad, Cornelius Aschermann, Lorenzo Fontana, Sasha Frolov, Ravi~Prakash Giri, Dhaval Kapil, Yiannis Kozyrakis, David LeBlanc, James Milazzo, Aleksandar Straumann, Gabriel Synnaeve, Varun Vontimitta, Spencer Whitman, and Joshua Saxe.
\newblock Purple llama cyberseceval: A secure coding benchmark for language models, 2023.
\newblock URL \url{https://arxiv.org/abs/2312.04724}.

\bibitem[Bianchi et~al.(2023)Bianchi, Suzgun, Attanasio, R{\"{o}}ttger, Jurafsky, Hashimoto, and Zou]{2023Federicoarxiv:2309.07875v2}
Federico Bianchi, Mirac Suzgun, Giuseppe Attanasio, Paul R{\"{o}}ttger, Dan Jurafsky, Tatsunori Hashimoto, and James Zou.
\newblock Safety-tuned llamas: Lessons from improving the safety of large language models that follow instructions.
\newblock \emph{CoRR}, abs/2309.07875, 2023.
\newblock URL \url{https://doi.org/10.48550/arXiv.2309.07875}.

\bibitem[Brown et~al.(2020)Brown, Mann, Ryder, Subbiah, Kaplan, Dhariwal, Neelakantan, Shyam, Sastry, Askell, et~al.]{brown2020language}
Tom Brown, Benjamin Mann, Nick Ryder, Melanie Subbiah, Jared~D Kaplan, Prafulla Dhariwal, Arvind Neelakantan, Pranav Shyam, Girish Sastry, Amanda Askell, et~al.
\newblock Language models are few-shot learners.
\newblock \emph{Advances in neural information processing systems}, 33:\penalty0 1877--1901, 2020.

\bibitem[Casper et~al.(2023)Casper, Davies, Shi, Gilbert, Scheurer, Rando, Freedman, Korbak, Lindner, Freire, et~al.]{casper2023open}
Stephen Casper, Xander Davies, Claudia Shi, Thomas~Krendl Gilbert, J{\'e}r{\'e}my Scheurer, Javier Rando, Rachel Freedman, Tomasz Korbak, David Lindner, Pedro Freire, et~al.
\newblock Open problems and fundamental limitations of reinforcement learning from human feedback.
\newblock \emph{arXiv preprint arXiv:2307.15217}, 2023.

\bibitem[Cheng et~al.(2025)Cheng, Hao, Liu, Zhou, Xie, Yao, Bian, Zhuang, Dey, Zha, et~al.]{cheng2025revisiting}
Zhoujun Cheng, Shibo Hao, Tianyang Liu, Fan Zhou, Yutao Xie, Feng Yao, Yuexin Bian, Yonghao Zhuang, Nilabjo Dey, Yuheng Zha, et~al.
\newblock Revisiting reinforcement learning for llm reasoning from a cross-domain perspective.
\newblock \emph{arXiv preprint arXiv:2506.14965}, 2025.

\bibitem[Cobbe et~al.(2021)Cobbe, Kosaraju, Bavarian, Chen, Jun, Kaiser, Plappert, Tworek, Hilton, Nakano, et~al.]{cobbe2021training}
Karl Cobbe, Vineet Kosaraju, Mohammad Bavarian, Mark Chen, Heewoo Jun, Lukasz Kaiser, Matthias Plappert, Jerry Tworek, Jacob Hilton, Reiichiro Nakano, et~al.
\newblock Training verifiers to solve math word problems.
\newblock \emph{arXiv preprint arXiv:2110.14168}, 2021.

\bibitem[DeepSeek(2025)]{deepseek_v31_release_2025}
DeepSeek.
\newblock Deepseek-v3.1 release, August 2025.
\newblock URL \url{https://api-docs.deepseek.com/news/news250821}.

\bibitem[Dinan et~al.(2019)Dinan, Humeau, Chintagunta, and Weston]{dinan-etal-2019-build}
Emily Dinan, Samuel Humeau, Bharath Chintagunta, and Jason Weston.
\newblock Build it break it fix it for dialogue safety: Robustness from adversarial human attack.
\newblock In Kentaro Inui, Jing Jiang, Vincent Ng, and Xiaojun Wan, editors, \emph{Proceedings of the 2019 Conference on Empirical Methods in Natural Language Processing and the 9th International Joint Conference on Natural Language Processing (EMNLP-IJCNLP)}, pages 4537--4546, Hong Kong, China, November 2019. Association for Computational Linguistics.
\newblock URL \url{https://aclanthology.org/D19-1461/}.

\bibitem[Dubey et~al.(2024)Dubey, Jauhri, Pandey, Kadian, Al-Dahle, Letman, Mathur, Schelten, Yang, Fan, et~al.]{llama3}
Abhimanyu Dubey, Abhinav Jauhri, Abhinav Pandey, Abhishek Kadian, Ahmad Al-Dahle, Aiesha Letman, Akhil Mathur, Alan Schelten, Amy Yang, Angela Fan, et~al.
\newblock The llama 3 herd of models.
\newblock \emph{arXiv e-prints}, pages arXiv--2407, 2024.

\bibitem[Evans(2010)]{evans2010intuition}
Jonathan St~BT Evans.
\newblock Intuition and reasoning: A dual-process perspective.
\newblock \emph{Psychological Inquiry}, 21\penalty0 (4):\penalty0 313--326, 2010.

\bibitem[Gao et~al.(2024)Gao, Song, Yang, Cai, Miao, Dong, Li, Ma, Chen, Xu, et~al.]{gao2024omni}
Bofei Gao, Feifan Song, Zhe Yang, Zefan Cai, Yibo Miao, Qingxiu Dong, Lei Li, Chenghao Ma, Liang Chen, Runxin Xu, et~al.
\newblock Omni-math: A universal olympiad level mathematic benchmark for large language models.
\newblock \emph{arXiv preprint arXiv:2410.07985}, 2024.

\bibitem[{Google DeepMind}(2025)]{GoogleDeepMind2025Gemini}
{Google DeepMind}.
\newblock {Gemini 2.5: Our newest Gemini model with thinking - The Keyword}.
\newblock \url{https://blog.google/technology/google-deepmind/gemini-model-thinking-updates-march-2025/#gemini-2-5-thinking}, March 2025.

\bibitem[Guha et~al.(2025)Guha, Marten, Keh, Raoof, Smyrnis, Bansal, Nezhurina, Mercat, Vu, Sprague, Suvarna, Feuer, Chen, Khan, Frankel, Grover, Choi, Muennighoff, Su, Zhao, Yang, Pimpalgaonkar, Sharma, Ji, Deng, Pratt, Ramanujan, Saad-Falcon, Li, Dave, Albalak, Arora, Wulfe, Hegde, Durrett, Oh, Bansal, Gabriel, Grover, Chang, Shankar, Gokaslan, Merrill, Hashimoto, Choi, Jitsev, Heckel, Sathiamoorthy, Dimakis, and Schmidt]{guha2025openthoughts}
Etash Guha, Ryan Marten, Sedrick Keh, Negin Raoof, Georgios Smyrnis, Hritik Bansal, Marianna Nezhurina, Jean Mercat, Trung Vu, Zayne Sprague, Ashima Suvarna, Benjamin Feuer, Liangyu Chen, Zaid Khan, Eric Frankel, Sachin Grover, Caroline Choi, Niklas Muennighoff, Shiye Su, Wanjia Zhao, John Yang, Shreyas Pimpalgaonkar, Kartik Sharma, Charlie Cheng-Jie Ji, Yichuan Deng, Sarah Pratt, Vivek Ramanujan, Jon Saad-Falcon, Jeffrey Li, Achal Dave, Alon Albalak, Kushal Arora, Blake Wulfe, Chinmay Hegde, Greg Durrett, Sewoong Oh, Mohit Bansal, Saadia Gabriel, Aditya Grover, Kai-Wei Chang, Vaishaal Shankar, Aaron Gokaslan, Mike~A. Merrill, Tatsunori Hashimoto, Yejin Choi, Jenia Jitsev, Reinhard Heckel, Maheswaran Sathiamoorthy, Alexandros~G. Dimakis, and Ludwig Schmidt.
\newblock Openthoughts: Data recipes for reasoning models.
\newblock \emph{arXiv preprint arXiv:2506.04178}, 2025.
\newblock URL \url{https://arxiv.org/abs/2506.04178}.

\bibitem[Gui et~al.(2025)Gui, Wang, Wang, Ma, Zhen, Yuan, HAO, Lian, Chen, and Wu]{gui2025hypertree}
Runquan Gui, Zhihai Wang, Jie Wang, Chi Ma, Huiling Zhen, Mingxuan Yuan, Jianye HAO, Defu Lian, Enhong Chen, and Feng Wu.
\newblock Hypertree planning: Enhancing llm reasoning via hierarchical thinking.
\newblock In \emph{Forty-second International Conference on Machine Learning}, 2025.

\bibitem[Guo et~al.(2025)Guo, Yang, Zhang, Song, Zhang, Xu, Zhu, Ma, Wang, Bi, et~al.]{guo2025deepseekr1}
Daya Guo, Dejian Yang, Haowei Zhang, Junxiao Song, Ruoyu Zhang, Runxin Xu, Qihao Zhu, Shirong Ma, Peiyi Wang, Xiao Bi, et~al.
\newblock Deepseek-r1: Incentivizing reasoning capability in llms via reinforcement learning.
\newblock \emph{arXiv preprint arXiv:2501.12948}, 2025.

\bibitem[Hartvigsen et~al.(2022)Hartvigsen, Gabriel, Palangi, Sap, Ray, and Kamar]{hartvigsen2022toxigen}
Thomas Hartvigsen, Saadia Gabriel, Hamid Palangi, Maarten Sap, Dipankar Ray, and Ece Kamar.
\newblock Toxigen: A large-scale machine-generated dataset for adversarial and implicit hate speech detection.
\newblock In \emph{Proceedings of the 60th Annual Meeting of the Association for Computational Linguistics (Volume 1: Long Papers)}, pages 3309--3326, 2022.

\bibitem[He et~al.(2025)He, Liu, Liu, Yan, Wang, Cheng, Zhang, Zhang, Xu, Shen, Li, Zeng, Wei, Cheng, An, Liu, and Zhou]{skywork-or1-2025}
Jujie He, Jiacai Liu, Chris~Yuhao Liu, Rui Yan, Chaojie Wang, Peng Cheng, Xiaoyu Zhang, Fuxiang Zhang, Jiacheng Xu, Wei Shen, Siyuan Li, Liang Zeng, Tianwen Wei, Cheng Cheng, Bo~An, Yang Liu, and Yahui Zhou.
\newblock Skywork open reasoner series.
\newblock \url{https://capricious-hydrogen-41c.notion.site/Skywork-Open-Reaonser-Series-1d0bc9ae823a80459b46c149e4f51680}, 2025.
\newblock Notion Blog.

\bibitem[Hosseini et~al.(2023)Hosseini, Palangi, and Awadallah]{hosseini2023empirical}
Saghar Hosseini, Hamid Palangi, and Ahmed~Hassan Awadallah.
\newblock An empirical study of metrics to measure representational harms in pre-trained language models.
\newblock \emph{CoRR}, abs/2301.09211, 2023.
\newblock URL \url{https://doi.org/10.48550/arXiv.2301.09211}.

\bibitem[Hu et~al.(2025)Hu, Zhang, Han, Jiang, Zhang, and Shum]{hu2025open}
Jingcheng Hu, Yinmin Zhang, Qi~Han, Daxin Jiang, Xiangyu Zhang, and Heung-Yeung Shum.
\newblock Open-reasoner-zero: An open source approach to scaling up reinforcement learning on the base model.
\newblock \emph{arXiv preprint arXiv:2503.24290}, 2025.

\bibitem[Hui et~al.(2024)Hui, Yang, Cui, Yang, Liu, Zhang, Liu, Zhang, Yu, Lu, et~al.]{qwen2.5code}
Binyuan Hui, Jian Yang, Zeyu Cui, Jiaxi Yang, Dayiheng Liu, Lei Zhang, Tianyu Liu, Jiajun Zhang, Bowen Yu, Keming Lu, et~al.
\newblock Qwen2. 5-coder technical report.
\newblock \emph{arXiv preprint arXiv:2409.12186}, 2024.

\bibitem[Jain et~al.(2024)Jain, Han, Gu, Li, Yan, Zhang, Wang, Solar-Lezama, Sen, and Stoica]{jain2024livecodebench}
Naman Jain, King Han, Alex Gu, Wen-Ding Li, Fanjia Yan, Tianjun Zhang, Sida Wang, Armando Solar-Lezama, Koushik Sen, and Ion Stoica.
\newblock Livecodebench: Holistic and contamination free evaluation of large language models for code.
\newblock \emph{arXiv preprint arXiv:2403.07974}, 2024.

\bibitem[Ji et~al.(2025{\natexlab{a}})Ji, Li, Ye, Wu, Yao, Xu, Mo, and Zhang]{ji2025test}
Yixin Ji, Juntao Li, Hai Ye, Kaixin Wu, Kai Yao, Jia Xu, Linjian Mo, and Min Zhang.
\newblock Test-time compute: from system-1 thinking to system-2 thinking.
\newblock \emph{arXiv preprint arXiv:2501.02497}, 2025{\natexlab{a}}.

\bibitem[Ji et~al.(2025{\natexlab{b}})Ji, Tian, Zhao, Wang, Chen, Peng, Zhao, and Li]{ji2025thinking}
Yunjie Ji, Xiaoyu Tian, Sitong Zhao, Haotian Wang, Shuaiting Chen, Yiping Peng, Han Zhao, and Xiangang Li.
\newblock Am-thinking-v1: Advancing the frontier of reasoning at 32b scale.
\newblock \emph{arXiv preprint arXiv:2505.08311}, 2025{\natexlab{b}}.

\bibitem[Kong et~al.(2023)Kong, Zhao, Chen, Li, Qin, Sun, Zhou, Wang, and Dong]{kong2023better}
Aobo Kong, Shiwan Zhao, Hao Chen, Qicheng Li, Yong Qin, Ruiqi Sun, Xin Zhou, Enzhi Wang, and Xiaohang Dong.
\newblock Better zero-shot reasoning with role-play prompting.
\newblock \emph{arXiv preprint arXiv:2308.07702}, 2023.

\bibitem[Leviathan et~al.(2023)Leviathan, Kalman, and Matias]{leviathan2023fast}
Yaniv Leviathan, Matan Kalman, and Yossi Matias.
\newblock Fast inference from transformers via speculative decoding.
\newblock In \emph{International Conference on Machine Learning}, pages 19274--19286. PMLR, 2023.

\bibitem[Li et~al.(2023)Li, Guo, Fan, Xu, Huang, Meng, and Song]{2023Haoranarxiv:2304.05197v3}
Haoran Li, Dadi Guo, Wei Fan, Mingshi Xu, Jie Huang, Fanpu Meng, and Yangqiu Song.
\newblock Multi-step jailbreaking privacy attacks on chatgpt.
\newblock In \emph{Findings of the Association for Computational Linguistics: {EMNLP} 2023, Singapore, December 6-10, 2023}, pages 4138--4153, 2023.
\newblock URL \url{https://aclanthology.org/2023.findings-emnlp.272}.

\bibitem[Lin et~al.(2024)Lin, Mu, Zhai, Wang, Wang, Wang, Gao, Zhang, Che, Baldwin, Han, and Li]{lin2024achillesheelsurveyred}
Lizhi Lin, Honglin Mu, Zenan Zhai, Minghan Wang, Yuxia Wang, Renxi Wang, Junjie Gao, Yixuan Zhang, Wanxiang Che, Timothy Baldwin, Xudong Han, and Haonan Li.
\newblock Against the achilles' heel: A survey on red teaming for generative models, 2024.
\newblock URL \url{https://arxiv.org/abs/2404.00629}.

\bibitem[Liu et~al.(2023{\natexlab{a}})Liu, Zhao, Qing, Kang, Sun, Kuang, and Wu]{liu2023Goal-Oriented}
Chengyuan Liu, Fubang Zhao, Lizhi Qing, Yangyang Kang, Changlong Sun, Kun Kuang, and Fei Wu.
\newblock Goal-oriented prompt attack and safety evaluation for llms, 2023{\natexlab{a}}.
\newblock URL \url{https://arxiv.org/abs/2309.11830}.

\bibitem[Liu et~al.(2025{\natexlab{a}})Liu, Diao, Lu, Hu, Dong, Choi, Kautz, and Dong]{prolongRL}
Mingjie Liu, Shizhe Diao, Ximing Lu, Jian Hu, Xin Dong, Yejin Choi, Jan Kautz, and Yi~Dong.
\newblock Prorl: Prolonged reinforcement learning expands reasoning boundaries in large language models, 2025{\natexlab{a}}.
\newblock URL \url{https://arxiv.org/abs/2505.24864}.

\bibitem[Liu et~al.(2023{\natexlab{b}})Liu, Deng, Li, Wang, Zhang, Liu, Wang, Zheng, and Liu]{liu2023promptinjection}
Yi~Liu, Gelei Deng, Yuekang Li, Kailong Wang, Tianwei Zhang, Yepang Liu, Haoyu Wang, Yan Zheng, and Yang Liu.
\newblock Prompt injection attack against llm-integrated applications, 2023{\natexlab{b}}.
\newblock URL \url{https://doi.org/10.48550/arXiv.2306.05499}.

\bibitem[Liu et~al.(2019)Liu, Ott, Goyal, Du, Joshi, Chen, Levy, Lewis, Zettlemoyer, and Stoyanov]{roberta}
Yinhan Liu, Myle Ott, Naman Goyal, Jingfei Du, Mandar Joshi, Danqi Chen, Omer Levy, Mike Lewis, Luke Zettlemoyer, and Veselin Stoyanov.
\newblock Roberta: A robustly optimized bert pretraining approach.
\newblock \emph{arXiv preprint arXiv:1907.11692}, 2019.

\bibitem[Liu et~al.(2025{\natexlab{b}})Liu, Yang, Chen, Lee, Shoeybi, Catanzaro, and Ping]{liu2025acereason}
Zihan Liu, Zhuolin Yang, Yang Chen, Chankyu Lee, Mohammad Shoeybi, Bryan Catanzaro, and Wei Ping.
\newblock Acereason-nemotron 1.1: Advancing math and code reasoning through sft and rl synergy.
\newblock \emph{arXiv preprint arXiv:2506.13284}, 2025{\natexlab{b}}.

\bibitem[Liu et~al.()Liu, Pang, Balabanov, Yang, Huang, Yin, Yang, and Liu]{liuliftreasoning}
Zihang Liu, Tianyu Pang, Oleg Balabanov, Chaoqun Yang, Tianjin Huang, Lu~Yin, Yaoqing Yang, and Shiwei Liu.
\newblock Lift the veil for the truth: Principal weights emerge after rank reduction for reasoning-focused supervised fine-tuning.
\newblock In \emph{Forty-second International Conference on Machine Learning}.

\bibitem[Luo et~al.(2025{\natexlab{a}})Luo, Tan, Huang, Patel, Ariyak, Wu, Shi, Xin, Cai, Weber, Zhang, Li, Popa, and Stoica]{deepcoder2025}
Michael Luo, Sijun Tan, Roy Huang, Ameen Patel, Alpay Ariyak, Qingyang Wu, Xiaoxiang Shi, Rachel Xin, Colin Cai, Maurice Weber, Ce~Zhang, Li~Erran Li, Raluca~Ada Popa, and Ion Stoica.
\newblock Deepcoder: A fully open-source 14b coder at o3-mini level, 2025{\natexlab{a}}.
\newblock URL \url{https://pretty-radio-b75.notion.site/DeepCoder-A-Fully-Open-Source-14B-Coder-at-O3-mini-Level-1cf81902c14680b3bee5eb349a512a51}.
\newblock Notion Blog.

\bibitem[Luo et~al.(2025{\natexlab{b}})Luo, Tan, Wong, Shi, Tang, Roongta, Cai, Luo, Li, Popa, and Stoica]{deepscaler2025}
Michael Luo, Sijun Tan, Justin Wong, Xiaoxiang Shi, William~Y. Tang, Manan Roongta, Colin Cai, Jeffrey Luo, Li~Erran Li, Raluca~Ada Popa, and Ion Stoica.
\newblock Deepscaler: Surpassing o1-preview with a 1.5b model by scaling rl, 2025{\natexlab{b}}.
\newblock URL \url{https://pretty-radio-b75.notion.site/DeepScaleR-Surpassing-O1-Preview-with-a-1-5B-Model-by-Scaling-RL-19681902c1468005bed8ca303013a4e2}.
\newblock Notion Blog.

\bibitem[Ma et~al.(2025)Ma, Liu, Jiang, Zhang, Ma, and Chen]{generalreasoner}
Xueguang Ma, Qian Liu, Dongfu Jiang, Ge~Zhang, Zejun Ma, and Wenhu Chen.
\newblock General-reasoner: Advancing llm reasoning across all domains.
\newblock \url{https://github.com/TIGER-AI-Lab/General-Reasoner/blob/main/General_Reasoner.pdf}, 2025.

\bibitem[MAA(2024)]{AIME2024}
MAA.
\newblock American invitational mathematics examination - aime.
\newblock In \emph{American Invitational Mathematics Examination - AIME 2024}, February 2024.
\newblock URL \url{https://maa.org/math-competitions/american-invitational-mathematics-examination-aime}.

\bibitem[Mazeika et~al.(2024)Mazeika, Phan, Yin, Zou, Wang, Mu, Sakhaee, Li, Basart, Li, Forsyth, and Hendrycks]{mazeika2024harmbench}
Mantas Mazeika, Long Phan, Xuwang Yin, Andy Zou, Zifan Wang, Norman Mu, Elham Sakhaee, Nathaniel Li, Steven Basart, Bo~Li, David~A. Forsyth, and Dan Hendrycks.
\newblock Harmbench: {A} standardized evaluation framework for automated red teaming and robust refusal.
\newblock \emph{CoRR}, abs/2402.04249, 2024.
\newblock URL \url{https://doi.org/10.48550/arXiv.2402.04249}.

\bibitem[Muennighoff et~al.(2025)Muennighoff, Yang, Shi, Li, Fei-Fei, Hajishirzi, Zettlemoyer, Liang, Cand{\`e}s, and Hashimoto]{s1}
Niklas Muennighoff, Zitong Yang, Weijia Shi, Xiang~Lisa Li, Li~Fei-Fei, Hannaneh Hajishirzi, Luke Zettlemoyer, Percy Liang, Emmanuel Cand{\`e}s, and Tatsunori Hashimoto.
\newblock s1: Simple test-time scaling.
\newblock \emph{arXiv preprint arXiv:2501.19393}, 2025.

\bibitem[Nakano et~al.(2021)Nakano, Hilton, Balaji, Wu, Ouyang, Kim, Hesse, Jain, Kosaraju, Saunders, Jiang, Cobbe, Eloundou, Krueger, Button, Knight, Chess, and Schulman]{nakano2021webgpt}
Reiichiro Nakano, Jacob Hilton, Suchir Balaji, Jeff Wu, Long Ouyang, Christina Kim, Christopher Hesse, Shantanu Jain, Vineet Kosaraju, William Saunders, Xu~Jiang, Karl Cobbe, Tyna Eloundou, Gretchen Krueger, Kevin Button, Matthew Knight, Benjamin Chess, and John Schulman.
\newblock Webgpt: Browser-assisted question-answering with human feedback.
\newblock \emph{arXiv preprint arXiv:2112.09332}, 2021.

\bibitem[NVIDIA(2025)]{nvidia2025openreasoning}
NVIDIA.
\newblock Openreasoning-nemotron-32b, july 2025.
\newblock URL \url{https://huggingface.co/nvidia/OpenReasoning-Nemotron-32B}.
\newblock Large language model for mathematical, coding, and scientific reasoning. Based on Qwen2.5-32B-Instruct with 32B parameters. Released July 16, 2025.

\bibitem[OpenAI(2024)]{OpenAIo1}
OpenAI.
\newblock {OpenAI o1 System Card}.
\newblock \url{https://openai.com/index/openai-o1-system-card/}, 2024.

\bibitem[{OpenAI}(2025)]{openai_gpt5_2025}
{OpenAI}.
\newblock {Introducing GPT-5}.
\newblock \url{https://openai.com/index/introducing-gpt-5/}, 2025.
\newblock Accessed: 2025-09-04.

\bibitem[OpenAI(2025)]{openaio3}
OpenAI.
\newblock Introducing openai o3 and o4-mini, 2025.
\newblock URL \url{https://openai.com/index/introducing-o3-and-o4-mini/}.
\newblock Accessed: 2025-06-12.

\bibitem[Ouyang et~al.(2022)Ouyang, Wu, Jiang, Almeida, Wainwright, Mishkin, Zhang, Agarwal, Slama, Ray, et~al.]{InstructGPT}
Long Ouyang, Jeffrey Wu, Xu~Jiang, Diogo Almeida, Carroll Wainwright, Pamela Mishkin, Chong Zhang, Sandhini Agarwal, Katarina Slama, Alex Ray, et~al.
\newblock Training language models to follow instructions with human feedback.
\newblock \emph{Advances in neural information processing systems}, 35:\penalty0 27730--27744, 2022.

\bibitem[Parmar et~al.(2025)Parmar, Liu, Goyal, Chen, Le, Mishra, Mobahi, Gu, Wang, Nakhost, et~al.]{parmar2025plangen}
Mihir Parmar, Xin Liu, Palash Goyal, Yanfei Chen, Long Le, Swaroop Mishra, Hossein Mobahi, Jindong Gu, Zifeng Wang, Hootan Nakhost, et~al.
\newblock Plangen: A multi-agent framework for generating planning and reasoning trajectories for complex problem solving.
\newblock \emph{arXiv preprint arXiv:2502.16111}, 2025.

\bibitem[Phan et~al.(2025)Phan, Gatti, Han, Li, Hu, Zhang, Zhang, Shaaban, Ling, Shi, et~al.]{phan2025humanity}
Long Phan, Alice Gatti, Ziwen Han, Nathaniel Li, Josephina Hu, Hugh Zhang, Chen Bo~Calvin Zhang, Mohamed Shaaban, John Ling, Sean Shi, et~al.
\newblock Humanity's last exam.
\newblock \emph{arXiv preprint arXiv:2501.14249}, 2025.

\bibitem[Qiu et~al.(2023)Qiu, Zhang, Li, He, and Lan]{2023Huachuanarxiv:2307.08487v3}
Huachuan Qiu, Shuai Zhang, Anqi Li, Hongliang He, and Zhenzhong Lan.
\newblock Latent jailbreak: {A} benchmark for evaluating text safety and output robustness of large language models.
\newblock \emph{CoRR}, abs/2307.08487, 2023.
\newblock URL \url{https://doi.org/10.48550/arXiv.2307.08487}.

\bibitem[Raffel et~al.(2020)Raffel, Shazeer, Roberts, Lee, Narang, Matena, Zhou, Li, and Liu]{raffel2020exploring}
Colin Raffel, Noam Shazeer, Adam Roberts, Katherine Lee, Sharan Narang, Michael Matena, Yanqi Zhou, Wei Li, and Peter~J Liu.
\newblock Exploring the limits of transfer learning with a unified text-to-text transformer.
\newblock \emph{Journal of machine learning research}, 21\penalty0 (140):\penalty0 1--67, 2020.

\bibitem[Rastogi et~al.(2025)Rastogi, Jiang, Lo, Berrada, Lample, Rute, Barmentlo, Yadav, Khandelwal, Chandu, et~al.]{rastogi2025magistral}
Abhinav Rastogi, Albert~Q Jiang, Andy Lo, Gabrielle Berrada, Guillaume Lample, Jason Rute, Joep Barmentlo, Karmesh Yadav, Kartik Khandelwal, Khyathi~Raghavi Chandu, et~al.
\newblock Magistral.
\newblock \emph{arXiv preprint arXiv:2506.10910}, 2025.

\bibitem[Rein et~al.(2023)Rein, Hou, Stickland, Petty, Pang, Dirani, Michael, and Bowman]{gpqa}
David Rein, Betty~Li Hou, Asa~Cooper Stickland, Jackson Petty, Richard~Yuanzhe Pang, Julien Dirani, Julian Michael, and Samuel~R. Bowman.
\newblock Gpqa: A graduate-level google-proof q\&a benchmark, 2023.
\newblock URL \url{https://arxiv.org/abs/2311.12022}.

\bibitem[Schulhoff et~al.(2023)Schulhoff, Pinto, Khan, Bouchard, Si, Boyd-Graber, Anati, Tagliabue, Kost, and Carnahan]{schulhoff-etal-2023-ignore}
Sander~V Schulhoff, Jeremy Pinto, Anaum Khan, Louis-FranÃois Bouchard, Chenglei Si, Jordan~Lee Boyd-Graber, Svetlina Anati, Valen Tagliabue, Anson~Liu Kost, and Christopher~R Carnahan.
\newblock Ignore this title and hackaprompt: Exposing systemic vulnerabilities of llms through a global prompt hacking competition.
\newblock In \emph{Empirical Methods in Natural Language Processing}, 2023.

\bibitem[Schuurmans et~al.(2024)Schuurmans, Dai, and Zanini]{schuurmans2024autoregressive}
Dale Schuurmans, Hanjun Dai, and Francesco Zanini.
\newblock Autoregressive large language models are computationally universal.
\newblock \emph{arXiv preprint arXiv:2410.03170}, 2024.

\bibitem[Shah et~al.(2023)Shah, Feuillade{-}Montixi, Pour, Tagade, Casper, and Rando]{Shah2023personamodulation}
Rusheb Shah, Quentin Feuillade{-}Montixi, Soroush Pour, Arush Tagade, Stephen Casper, and Javier Rando.
\newblock Scalable and transferable black-box jailbreaks for language models via persona modulation.
\newblock \emph{CoRR}, abs/2311.03348, 2023.
\newblock URL \url{https://doi.org/10.48550/arXiv.2311.03348}.

\bibitem[Shaikh et~al.(2023)Shaikh, Zhang, Held, Bernstein, and Yang]{shaikh2023second}
Omar Shaikh, Hongxin Zhang, William Held, Michael~S. Bernstein, and Diyi Yang.
\newblock On second thought, let's not think step by step! bias and toxicity in zero-shot reasoning.
\newblock In \emph{Proceedings of the 61st Annual Meeting of the Association for Computational Linguistics (Volume 1: Long Papers), {ACL} 2023, Toronto, Canada, July 9-14, 2023}, pages 4454--4470, 2023.
\newblock URL \url{https://doi.org/10.18653/v1/2023.acl-long.244}.

\bibitem[Shao et~al.(2025)Shao, Li, Xin, Geng, Wang, Oh, Du, Lambert, Min, Krishna, Tsvetkov, Hajishirzi, Koh, and Zettlemoyer]{shao2025spurious}
Rulin Shao, Shuyue~Stella Li, Rui Xin, Scott Geng, Yiping Wang, Sewoong Oh, Simon~Shaolei Du, Nathan Lambert, Sewon Min, Ranjay Krishna, Yulia Tsvetkov, Hannaneh Hajishirzi, Pang~Wei Koh, and Luke Zettlemoyer.
\newblock Spurious rewards: Rethinking training signals in rlvr.
\newblock \url{https://rethink-rlvr.notion.site/Spurious-Rewards-Rethinking-Training-Signals-in-RLVR-1f4df34dac1880948858f95aeb88872f}, 2025.
\newblock Notion Blog.

\bibitem[Shao et~al.(2024)Shao, Wang, Zhu, Xu, Song, Bi, Zhang, Zhang, Li, Wu, et~al.]{shao2024deepseekmath}
Zhihong Shao, Peiyi Wang, Qihao Zhu, Runxin Xu, Junxiao Song, Xiao Bi, Haowei Zhang, Mingchuan Zhang, YK~Li, Y~Wu, et~al.
\newblock Deepseekmath: Pushing the limits of mathematical reasoning in open language models.
\newblock \emph{arXiv preprint arXiv:2402.03300}, 2024.

\bibitem[Sharma(2024)]{optillm}
Asankhaya Sharma.
\newblock Optillm: Optimizing inference proxy for llms, 2024.
\newblock URL \url{https://github.com/codelion/optillm}.

\bibitem[Shen et~al.(2023)Shen, Chen, Backes, Shen, and Zhang]{shen2023do}
Xinyue Shen, Zeyuan Chen, Michael Backes, Yun Shen, and Yang Zhang.
\newblock "do anything now": Characterizing and evaluating in-the-wild jailbreak prompts on large language models.
\newblock \emph{CoRR}, abs/2308.03825, 2023.
\newblock URL \url{https://doi.org/10.48550/arXiv.2308.03825}.

\bibitem[Sheng et~al.(2025)Sheng, Zhang, Ye, Wu, Zhang, Zhang, Peng, Lin, and Wu]{sheng2025hybridflow}
Guangming Sheng, Chi Zhang, Zilingfeng Ye, Xibin Wu, Wang Zhang, Ru~Zhang, Yanghua Peng, Haibin Lin, and Chuan Wu.
\newblock Hybridflow: A flexible and efficient rlhf framework.
\newblock In \emph{Proceedings of the Twentieth European Conference on Computer Systems}, pages 1279--1297, 2025.

\bibitem[Shinn et~al.(2023)Shinn, Cassano, Gopinath, Narasimhan, and Yao]{shinn2023reflexion}
Noah Shinn, Federico Cassano, Ashwin Gopinath, Karthik Narasimhan, and Shunyu Yao.
\newblock Reflexion: Language agents with verbal reinforcement learning.
\newblock \emph{Advances in Neural Information Processing Systems}, 36:\penalty0 8634--8652, 2023.

\bibitem[Snell et~al.(2025)Snell, Lee, Xu, and Kumar]{snell2025scaling}
Charlie~Victor Snell, Jaehoon Lee, Kelvin Xu, and Aviral Kumar.
\newblock Scaling {LLM} test-time compute optimally can be more effective than scaling parameters for reasoning.
\newblock In \emph{The Thirteenth International Conference on Learning Representations}, 2025.
\newblock URL \url{https://openreview.net/forum?id=4FWAwZtd2n}.

\bibitem[Stiennon et~al.(2020)Stiennon, Ouyang, Wu, Ziegler, Lowe, Voss, Radford, Amodei, and Christiano]{stiennon2020learning}
Nisan Stiennon, Long Ouyang, Jeffrey Wu, Daniel Ziegler, Ryan Lowe, Chelsea Voss, Alec Radford, Dario Amodei, and Paul~F Christiano.
\newblock Learning to summarize with human feedback.
\newblock \emph{Advances in neural information processing systems}, 33:\penalty0 3008--3021, 2020.

\bibitem[Taori et~al.(2023)Taori, Gulrajani, Zhang, Dubois, Li, Guestrin, Liang, and Hashimoto]{alpaca}
Rohan Taori, Ishaan Gulrajani, Tianyi Zhang, Yann Dubois, Xuechen Li, Carlos Guestrin, Percy Liang, and Tatsunori~B Hashimoto.
\newblock Stanford alpaca: An instruction-following llama model, 2023.

\bibitem[Team(2025)]{team2025qwq}
Qwen Team.
\newblock Qwq-32b: Embracing the power of reinforcement learning, 2025.

\bibitem[Tian et~al.(2024)Tian, Gao, Zhang, Chen, Fan, Guo, Haas, Ji, Krongchon, Li, Liu, Luo, Ma, Tong, Trinh, Tian, Wang, Wu, Xiong, Yin, Zhu, Lieret, Lu, Liu, Du, Tao, Press, Callan, Huerta, and Peng]{tian2024scicode}
Minyang Tian, Luyu Gao, Shizhuo~Dylan Zhang, Xinan Chen, Cunwei Fan, Xuefei Guo, Roland Haas, Pan Ji, Kittithat Krongchon, Yao Li, Shengyan Liu, Di~Luo, Yutao Ma, Hao Tong, Kha Trinh, Chenyu Tian, Zihan Wang, Bohao Wu, Yanyu Xiong, Shengzhu Yin, Minhui Zhu, Kilian Lieret, Yanxin Lu, Genglin Liu, Yufeng Du, Tianhua Tao, Ofir Press, Jamie Callan, Eliu Huerta, and Hao Peng.
\newblock Scicode: A research coding benchmark curated by scientists, 2024.

\bibitem[Tian et~al.(2025)Tian, Ji, Wang, Chen, Zhao, Peng, Zhao, and Li]{tian2025not}
Xiaoyu Tian, Yunjie Ji, Haotian Wang, Shuaiting Chen, Sitong Zhao, Yiping Peng, Han Zhao, and Xiangang Li.
\newblock Not all correct answers are equal: Why your distillation source matters.
\newblock \emph{arXiv preprint arXiv:2505.14464}, 2025.
\newblock URL \url{https://arxiv.org/abs/2505.14464}.

\bibitem[Toyer et~al.(2023)Toyer, Watkins, Mendes, Svegliato, Bailey, Wang, Ong, Elmaaroufi, Abbeel, Darrell, Ritter, and Russell]{2023Samarxiv:2311.01011v1}
Sam Toyer, Olivia Watkins, Ethan~Adrian Mendes, Justin Svegliato, Luke Bailey, Tiffany Wang, Isaac Ong, Karim Elmaaroufi, Pieter Abbeel, Trevor Darrell, Alan Ritter, and Stuart Russell.
\newblock Tensor trust: Interpretable prompt injection attacks from an online game.
\newblock \emph{CoRR}, abs/2311.01011, 2023.
\newblock URL \url{https://doi.org/10.48550/arXiv.2311.01011}.

\bibitem[Vidgen et~al.(2023)Vidgen, Kirk, Qian, Scherrer, Kannappan, Hale, and R{\"{o}}ttger]{vidgen2023simplesafetytests}
Bertie Vidgen, Hannah~Rose Kirk, Rebecca Qian, Nino Scherrer, Anand Kannappan, Scott~A. Hale, and Paul R{\"{o}}ttger.
\newblock Simplesafetytests: a test suite for identifying critical safety risks in large language models.
\newblock \emph{CoRR}, abs/2311.08370, 2023.
\newblock URL \url{https://doi.org/10.48550/arXiv.2311.08370}.

\bibitem[Wang et~al.(2024{\natexlab{a}})Wang, Ye, Fang, and Li]{wang2024coplanner}
Danqing Wang, Zhuorui Ye, Fei Fang, and Lei Li.
\newblock Cooperative strategic planning enhances reasoning capabilities in large language models.
\newblock \emph{arXiv preprint arXiv:2410.20007}, 2024{\natexlab{a}}.

\bibitem[Wang et~al.(2025{\natexlab{a}})Wang, Li, Sun, Chen, Liu, Wu, Lu, Song, and Yadkori]{wang2025hrm}
Guan Wang, Jin Li, Yuhao Sun, Xing Chen, Changling Liu, Yue Wu, Meng Lu, Sen Song, and Yasin~Abbasi Yadkori.
\newblock Hierarchical reasoning model.
\newblock \emph{arXiv preprint arXiv:2506.21734}, 2025{\natexlab{a}}.

\bibitem[Wang et~al.(2024{\natexlab{b}})Wang, Wang, Athiwaratkun, Zhang, and Zou]{wang2024mixture}
Junlin Wang, Jue Wang, Ben Athiwaratkun, Ce~Zhang, and James Zou.
\newblock Mixture-of-agents enhances large language model capabilities.
\newblock \emph{arXiv preprint arXiv:2406.04692}, 2024{\natexlab{b}}.

\bibitem[Wang et~al.(2023{\natexlab{a}})Wang, Tu, Chen, Yuan, Huang, Jiao, and Lyu]{2023Wenxuanarxiv:2310.00905v1}
Wenxuan Wang, Zhaopeng Tu, Chang Chen, Youliang Yuan, Jen{-}tse Huang, Wenxiang Jiao, and Michael~R. Lyu.
\newblock All languages matter: On the multilingual safety of large language models.
\newblock \emph{CoRR}, abs/2310.00905, 2023{\natexlab{a}}.
\newblock URL \url{https://doi.org/10.48550/arXiv.2310.00905}.

\bibitem[Wang et~al.(2023{\natexlab{b}})Wang, Wei, Schuurmans, Le, Chi, Narang, Chowdhery, and Zhou]{wang2023selfconsistency}
Xuezhi Wang, Jason Wei, Dale Schuurmans, Quoc~V Le, Ed~H. Chi, Sharan Narang, Aakanksha Chowdhery, and Denny Zhou.
\newblock Self-consistency improves chain of thought reasoning in language models.
\newblock In \emph{The Eleventh International Conference on Learning Representations}, 2023{\natexlab{b}}.
\newblock URL \url{https://openreview.net/forum?id=1PL1NIMMrw}.

\bibitem[Wang et~al.(2025{\natexlab{b}})Wang, Yang, Zeng, Ren, Liu, Peng, Cheng, He, Wang, Gao, Chen, Wang, Du, and Shen]{one-shot-rl}
Yiping Wang, Qing Yang, Zhiyuan Zeng, Liliang Ren, Liyuan Liu, Baolin Peng, Hao Cheng, Xuehai He, Kuan Wang, Jianfeng Gao, Weizhu Chen, Shuohang Wang, Simon~Shaolei Du, and Yelong Shen.
\newblock Reinforcement learning for reasoning in large language models with one training example, 2025{\natexlab{b}}.
\newblock URL \url{https://arxiv.org/abs/2504.20571}.

\bibitem[Wang et~al.(2023{\natexlab{c}})Wang, Li, Han, Nakov, and Baldwin]{wang2023donotanswer}
Yuxia Wang, Haonan Li, Xudong Han, Preslav Nakov, and Timothy Baldwin.
\newblock Do-not-answer: {A} dataset for evaluating safeguards in llms.
\newblock \emph{CoRR}, abs/2308.13387, 2023{\natexlab{c}}.
\newblock URL \url{https://doi.org/10.48550/arXiv.2308.13387}.

\bibitem[Wang et~al.(2025{\natexlab{c}})Wang, Zhou, Li, and Liu]{wang2025octothinker}
Zengzhi Wang, Fan Zhou, Xuefeng Li, and Pengfei Liu.
\newblock Octothinker: Revisiting mid-training in the era of rl scaling.
\newblock \url{https://tinyurl.com/OctoThinker}, 2025{\natexlab{c}}.
\newblock Notion Blog.

\bibitem[Wei et~al.(2023{\natexlab{a}})Wei, Haghtalab, and Steinhardt]{wei2023jailbroken}
Alexander Wei, Nika Haghtalab, and Jacob Steinhardt.
\newblock Jailbroken: How does {LLM} safety training fail?
\newblock In \emph{Advances in Neural Information Processing Systems 36: Annual Conference on Neural Information Processing Systems 2023, NeurIPS 2023, New Orleans, LA, USA, December 10 - 16, 2023}, 2023{\natexlab{a}}.
\newblock URL \url{http://papers.nips.cc/paper\\_files/paper/2023/hash/fd6613131889a4b656206c50a8bd7790-Abstract-Conference.html}.

\bibitem[Wei et~al.(2021)Wei, Bosma, Zhao, Guu, Yu, Lester, Du, Dai, and Le]{flan}
Jason Wei, Maarten Bosma, Vincent~Y Zhao, Kelvin Guu, Adams~Wei Yu, Brian Lester, Nan Du, Andrew~M Dai, and Quoc~V Le.
\newblock Finetuned language models are zero-shot learners.
\newblock \emph{arXiv preprint arXiv:2109.01652}, 2021.

\bibitem[Wei et~al.(2022)Wei, Wang, Schuurmans, Bosma, Xia, Chi, Le, Zhou, et~al.]{wei2022chain}
Jason Wei, Xuezhi Wang, Dale Schuurmans, Maarten Bosma, Fei Xia, Ed~Chi, Quoc~V Le, Denny Zhou, et~al.
\newblock Chain-of-thought prompting elicits reasoning in large language models.
\newblock \emph{Advances in neural information processing systems}, 35:\penalty0 24824--24837, 2022.

\bibitem[Wei et~al.(2023{\natexlab{b}})Wei, Wang, and Wang]{2023Zemingarxiv:2310.06387v1}
Zeming Wei, Yifei Wang, and Yisen Wang.
\newblock Jailbreak and guard aligned language models with only few in-context demonstrations.
\newblock \emph{CoRR}, abs/2310.06387, 2023{\natexlab{b}}.
\newblock URL \url{https://doi.org/10.48550/arXiv.2310.06387}.

\bibitem[{xAI}(2025)]{xAI2025Grok4}
{xAI}.
\newblock {Grok 4}.
\newblock \url{https://x.ai/news/grok-4}, July 2025.

\bibitem[Xu et~al.(2024)Xu, Jin, Hao, Song, Sun, and Yuan]{xu2024llavacot}
Guowei Xu, Peng Jin, Li~Hao, Yibing Song, Lichao Sun, and Li~Yuan.
\newblock Llava-o1: Let vision language models reason step-by-step.
\newblock \emph{arXiv preprint arXiv:2411.10440}, 2024.

\bibitem[Yang et~al.(2024{\natexlab{a}})Yang, Yang, Zhang, Hui, Zheng, Yu, Li, Liu, Huang, Wei, et~al.]{yang2024qwen2}
An~Yang, Baosong Yang, Beichen Zhang, Binyuan Hui, Bo~Zheng, Bowen Yu, Chengyuan Li, Dayiheng Liu, Fei Huang, Haoran Wei, et~al.
\newblock Qwen2. 5 technical report.
\newblock \emph{arXiv preprint arXiv:2412.15115}, 2024{\natexlab{a}}.

\bibitem[Yang et~al.(2024{\natexlab{b}})Yang, Zhang, Hui, Gao, Yu, Li, Liu, Tu, Zhou, Lin, et~al.]{qwen2.5math}
An~Yang, Beichen Zhang, Binyuan Hui, Bofei Gao, Bowen Yu, Chengpeng Li, Dayiheng Liu, Jianhong Tu, Jingren Zhou, Junyang Lin, et~al.
\newblock Qwen2. 5-math technical report: Toward mathematical expert model via self-improvement.
\newblock \emph{arXiv preprint arXiv:2409.12122}, 2024{\natexlab{b}}.

\bibitem[Yang et~al.(2025{\natexlab{a}})Yang, Li, Yang, Zhang, Hui, Zheng, Yu, Gao, Huang, Lv, et~al.]{yang2025qwen3}
An~Yang, Anfeng Li, Baosong Yang, Beichen Zhang, Binyuan Hui, Bo~Zheng, Bowen Yu, Chang Gao, Chengen Huang, Chenxu Lv, et~al.
\newblock Qwen3 technical report.
\newblock \emph{arXiv preprint arXiv:2505.09388}, 2025{\natexlab{a}}.

\bibitem[Yang et~al.(2025{\natexlab{b}})Yang, Ma, Lin, and Wei]{yang2025towards}
Wenkai Yang, Shuming Ma, Yankai Lin, and Furu Wei.
\newblock Towards thinking-optimal scaling of test-time compute for llm reasoning.
\newblock \emph{arXiv preprint arXiv:2502.18080}, 2025{\natexlab{b}}.

\bibitem[Ye et~al.(2025{\natexlab{a}})Ye, Huang, Xiao, Chern, Xia, and Liu]{limo}
Yixin Ye, Zhen Huang, Yang Xiao, Ethan Chern, Shijie Xia, and Pengfei Liu.
\newblock Limo: Less is more for reasoning.
\newblock \emph{arXiv preprint arXiv:2502.03387}, 2025{\natexlab{a}}.

\bibitem[Ye et~al.(2025{\natexlab{b}})Ye, Xiao, Mi, and Liu]{AIME2025}
Yixin Ye, Yang Xiao, Tiantian Mi, and Pengfei Liu.
\newblock Aime-preview: A rigorous and immediate evaluation framework for advanced mathematical reasoning, 2025{\natexlab{b}}.

\bibitem[Yu et~al.(2025)Yu, Zhang, Zhu, Yuan, Zuo, Yue, Fan, Liu, Liu, Liu, et~al.]{yu2025dapo}
Qiying Yu, Zheng Zhang, Ruofei Zhu, Yufeng Yuan, Xiaochen Zuo, Yu~Yue, Tiantian Fan, Gaohong Liu, Lingjun Liu, Xin Liu, et~al.
\newblock Dapo: An open-source llm reinforcement learning system at scale.
\newblock \emph{arXiv preprint arXiv:2503.14476}, 2025.

\bibitem[Yue et~al.(2025)Yue, Chen, Lu, Zhao, Wang, Song, and Huang]{yue2025does}
Yang Yue, Zhiqi Chen, Rui Lu, Andrew Zhao, Zhaokai Wang, Shiji Song, and Gao Huang.
\newblock Does reinforcement learning really incentivize reasoning capacity in llms beyond the base model?
\newblock \emph{arXiv preprint arXiv:2504.13837}, 2025.

\bibitem[Zeng et~al.(2025)Zeng, Huang, Liu, Liu, He, Ma, and He]{zeng2025simplerl}
Weihao Zeng, Yuzhen Huang, Qian Liu, Wei Liu, Keqing He, Zejun Ma, and Junxian He.
\newblock Simplerl-zoo: Investigating and taming zero reinforcement learning for open base models in the wild.
\newblock \emph{arXiv preprint arXiv:2503.18892}, 2025.

\bibitem[Zhao et~al.(2025)Zhao, Kang, Feng, Levine, and Song]{zhao2025learning}
Xuandong Zhao, Zhewei Kang, Aosong Feng, Sergey Levine, and Dawn Song.
\newblock Learning to reason without external rewards.
\newblock \emph{arXiv preprint arXiv:2505.19590}, 2025.

\end{thebibliography}


\end{document}